\documentclass[a4paper,conference]{IEEEtran}
\usepackage{times}
\usepackage{epsfig}
\usepackage{graphicx}
\usepackage{amsmath}
\usepackage{amssymb}
\usepackage{subfigure}
\usepackage{multirow}
\usepackage{bbding}
\usepackage{caption,subcaption}
\usepackage{dblfloatfix}
\usepackage[normalem]{ulem}
\useunder{\uline}{\ul}{}

\makeatletter
\renewcommand{\@thesubfigure}{\hskip\subfiglabelskip}
\makeatother

\makeatletter
\@namedef{ver@everyshi.sty}{}
\makeatother
\usepackage{copyrightbox}
\usepackage{tikz}

\usepackage[pagebackref=true,breaklinks=true,colorlinks,bookmarks=false]{hyperref}

\ifCLASSINFOpdf
\else
\fi
\begin{document}
%
\title{PMPNet: Pixel Movement Prediction Network for Monocular Depth Estimation in Dynamic Scenes}

\author{\IEEEauthorblockN{Kebin Peng, John Quarles, Kevin Desai}
\IEEEauthorblockA{\{kebin.peng, john.quarles, kevin.desai\}@utsa.edu \\
Department of Computer Science\\
The University of Texas at San Antonio\\
One UTSA Circle, San Antonio, TX 78249
}
}


%


\maketitle

\begin{abstract}
In this paper, we propose a novel method for monocular depth estimation in dynamic scenes.
We first explore the arbitrariness of object's movement trajectory in dynamic scenes theoretically.
To overcome the arbitrariness, we use assume that points move along a straight line over short distances and then summarize it as a triangular constraint loss in two dimensional Euclidean space.
This triangular loss function is used as part of our proposed pixel movement prediction network, PMPNet, to estimate a dense depth map from a single input image.
To overcome the depth inconsistency problem around the edges, we propose a deformable support window module that learns features from different shapes of objects, making depth value more accurate around edge area. 
The proposed model is trained and tested on two outdoor datasets - KITTI and Make3D, as well as an indoor dataset - NYU Depth V2. The quantitative and qualitative results reported on these datasets demonstrate the success of our proposed model when compared against other approaches.
Ablation study results on the KITTI dataset also validate the effectiveness of the proposed pixel movement prediction module as well as the deformable support window module.
\end{abstract}


%
\IEEEpeerreviewmaketitle

\section{Introduction}\label{intro}


Monocular depth estimation for dynamic scenes, i.e., estimating depth value for a moving point seen from a monocular moving camera, is a challenging task. 
In static scenes, like Middleburry in the 2014 dataset \cite{scharstein2014high}, only camera moves and the objects are static. 
However in dynamic scenes, such as in autonomous driving, not only will the camera move but also the objects. The moving objects such as cars and trucks violate the static
world assumption \cite{klingner2020selfsupervised}.
Hence, the object motion relative to the camera should contain the movement of the object as well as the movement of the camera \cite{li2020unsupervised}. 
For this reason, previous works have used two neural networks, one for estimating depth the other one for estimating camera motion from monocular video \cite{zhou2017unsupervised} or synchronized stereo pairs \cite{garg2016unsupervised,godard2017unsupervised}. 
Alternatively, \cite{li2020unsupervised} used a neural network to recognize possible moving objects then estimate depth with other neural networks.

Although existing monocular depth estimation methods \cite{bhat2021adabins,song2021monocular,ranftl2021vision,wang2020sdc} perform well in autonomous driving datasets such as KITTI \cite{geiger2012we}, most of them do not consider dynamic scenes, which is common in real-world autonomous driving 
The methods that work on such dynamic scenes \cite{li2020unsupervised,garg2016unsupervised,godard2017unsupervised,di2019monocular,yoon2020novel} do not consider the arbitrariness of object's movement trajectory.
In other words, a pixel could move to any place in the next time instance and these methods do not add any constraints on the pixel/object movement trajectory.
Such arbitrariness could result in a huge search space, which makes it hard for neural networks to learn useful information.

\begin{figure}
  \begin{center}
	\subfigure[]
	{
	   \includegraphics[width=0.47\linewidth]{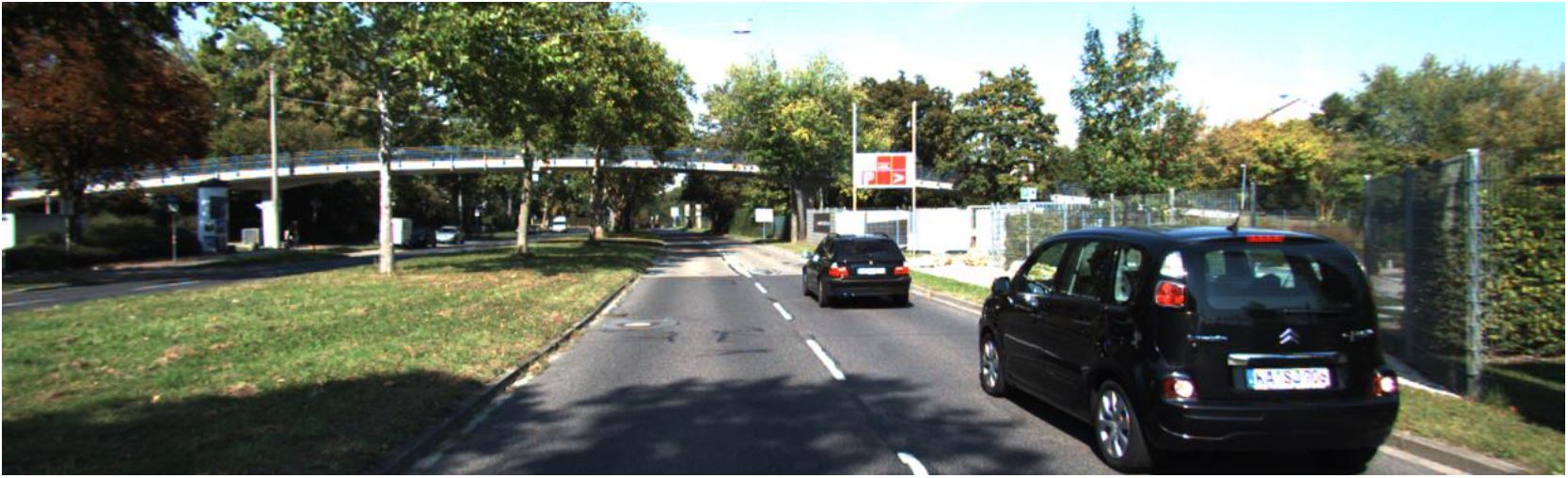}
	}\hspace{-0.2in}
	\quad
	\subfigure[] 
	{
		\includegraphics[width=0.47\linewidth]{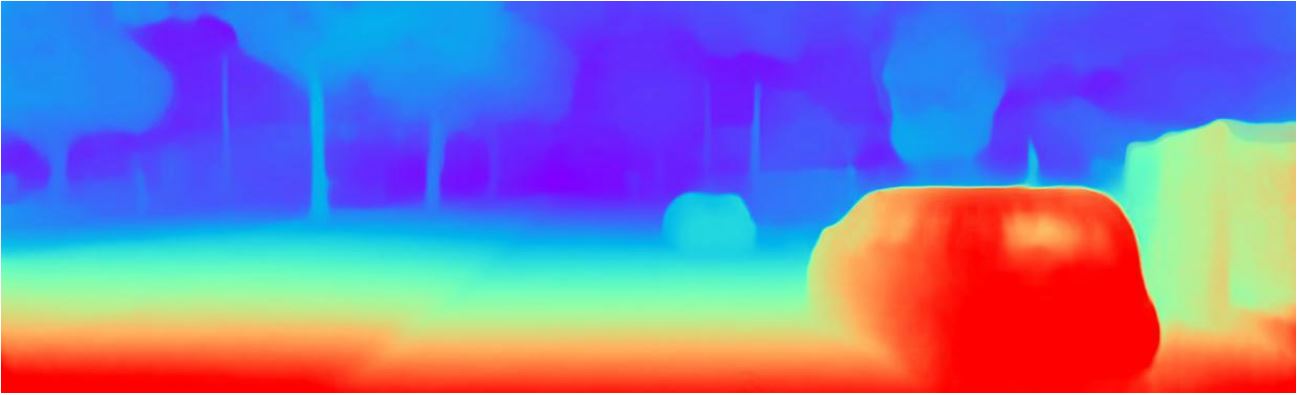}
	}\hspace{-0.2in}
	\quad
	\vspace{-7mm}
	
	\subfigure[]
	{ 
		\includegraphics[width=0.47\linewidth]{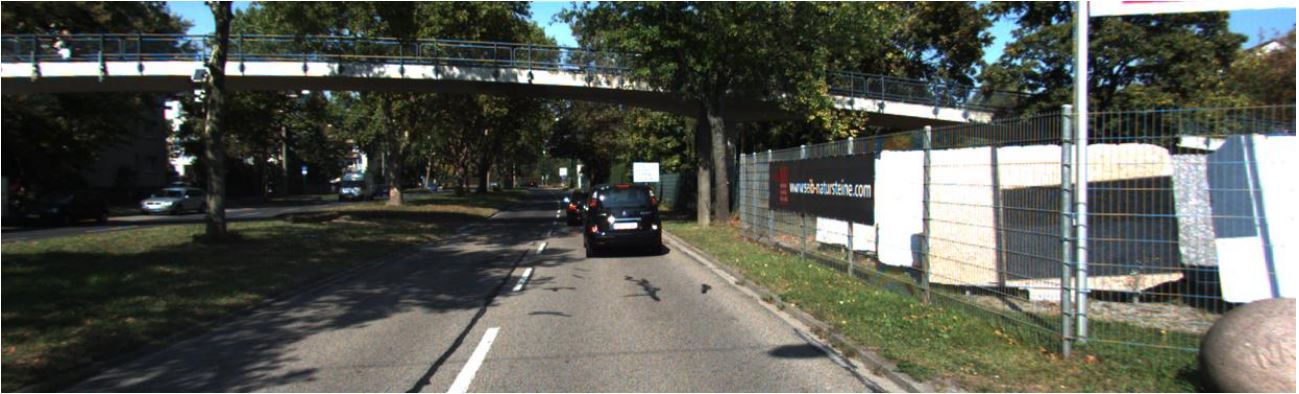}
	}\hspace{-0.2in}
	\quad
	\subfigure[] 
	{
		\includegraphics[width=0.47\linewidth]{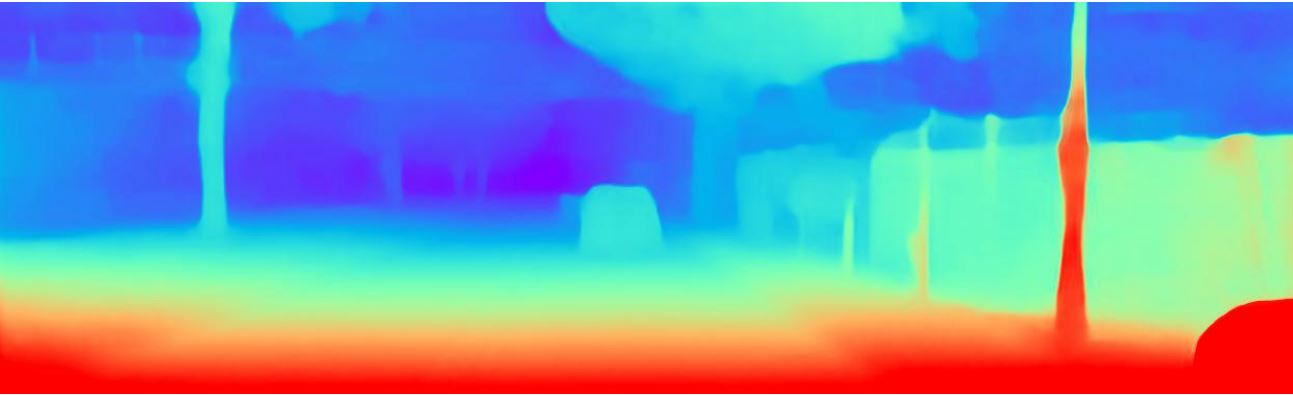} 
	}\hspace{-0.2in}
	\quad
    \vspace{-7mm}
    
	\subfigure[]
	{ 
		\includegraphics[width=0.47\linewidth]{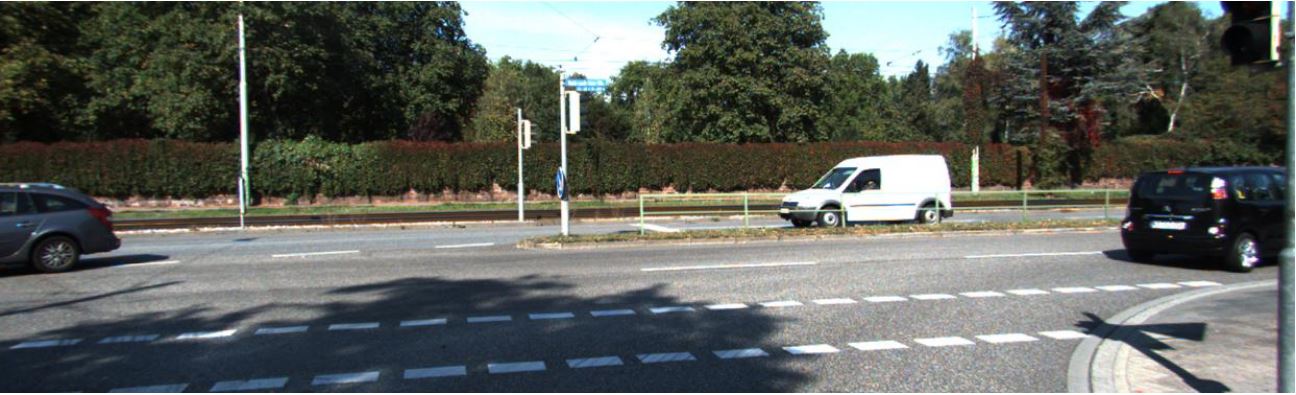}
	}\hspace{-0.2in}
	\quad
	\subfigure[] 
	{
		\includegraphics[width=0.47\linewidth]{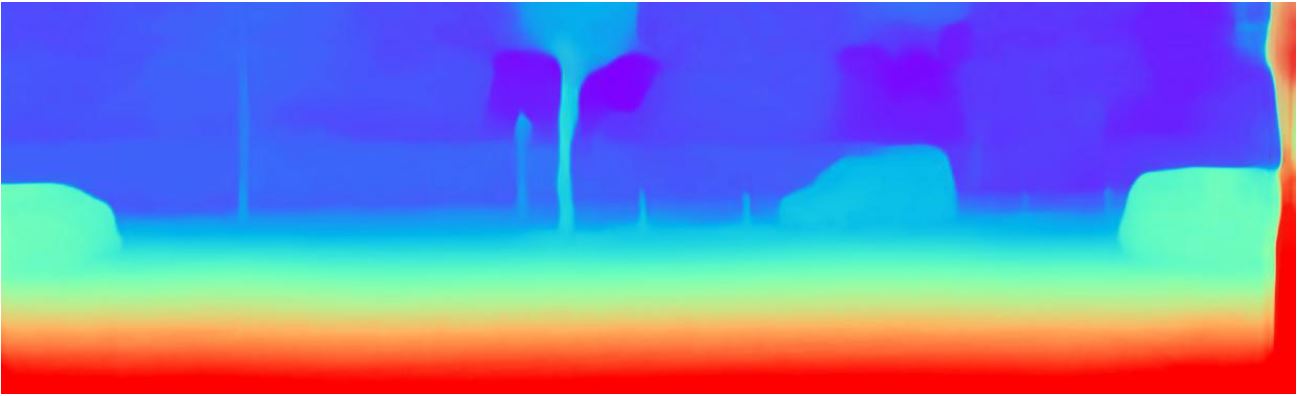} 
	}\hspace{-0.2in}
	\quad
	\vspace{-7mm}
	
	\caption{\textbf{Monocular Depth Estimation Results:} Two examples from the KITTI dataset \cite{geiger2012we}. Our model estimates accurate depth maps with non-blur object edges.}
	\label{KITTI example}
	\vspace{-8mm}
	\end{center} 
\end{figure}

In this paper, we try to reduce such arbitrariness by constraining the trajectory of the pixel.
For this goal, we use the assumption stated in \cite{avidan2000trajectory}, i.e., the pixel moving along a straight-line and the camera center of projection traces a straight line during camera motion.
Such an assumption is reasonable in an autonomous driving dataset because in most cases the cars and the camera on cars will move along a straight-line (left/right turn can be viewed as a combination of many small straight-lines).
Considering this assumption, we propose the \textit{Pixel Movement Prediction Module} to estimate two two-dimensional vectors $v_1$, $v_2$ that could span a null space.
As concluded by \cite{avidan2000trajectory}, in the null space, the straight-line $L$ could be represented by the linear combination of $v_1$, $v_2$.
We summarize this conclusion into a \textit{Pixel Movement Triangular Constrain Loss Function} to train our model.
Using the \textit{Pixel Movement Prediction Module}, we synchronize a ``right image" feature map according to the input image's feature map (the input image is viewed as the ``left image" of the stereo pairs).
One thing needs to be noticed is that we synchronize the feature map of ``right image" instead of the ``right image" itself. This is because the feature map can be used directly to construct the cost volume (see Section \ref{Method} - Formula \ref{eq:cost volume}).


In most monocular depth estimation methods, there exists depth inconsistency around edge area.
Such inconsistencies exist because standard CNNs consist of fixed geometric structures in their architecture, which restrict the learning to standard geometric transformations.
\cite{dai2017deformable} further proposed deformable convolutions that enable free form transformations of the sampling grid and thereby achieving higher accuracy.
In our work, we adopt deformable convolution layers which especially improves the depth estimation at the object edges, where the pixel movement may involve compound transformations.
These deformable convolution layers help better estimate depth near the edges.
Figure \ref{KITTI example} shows the depth estimation results of our proposed PMPNet on three different examples from the KITTI dataset \cite{geiger2012we}.

\subsection{Contributions}
Following are the major contributions of our work: 
\begin{itemize}
	\item We analyze the arbitrariness of object's / pixel's trajectory for dynamic scenes theoretically and use a straight-line assumption to constrain the trajectory. 
	\item  We propose a novel deep neural network architecture called PMPNet. It consists of a \textit{Pixel Movement Prediction Module} which provides two pixel movement predictions and a third straight-line prediction. The relation between pixel movements and straight-line is summarized into a novel \textit{Triangular Constraint Loss Function}.
	\item We propose a \textit{Deformable Convolution Support Window}, which better addresses the problem of blurred edges during monocular depth estimation.
	\vspace{-2mm}
\end{itemize}

\section{Related work}\label{related work}





\subsection{Supervised Monocular Depth Estimation}\label{sub-supervise}
\cite{bhat2021adabins} view Monocular Depth Estimation as a regression problem and proposed a transformer based architecture. It divides the depth range into bins whose center value is estimated adaptively per image.
\cite{ranftl2021vision} proposed using the vision transformers as a backbone for monocular depth estimation. 
\cite{ranjan2019competitive} proposed a framework where networks act as competitors and collaborators to reach specific goals by benefiting from each other.
\cite{fu2018deep} proposed the deep ordinal regression network, using spacing-increasing discretization strategy and the ordinary regression loss.
\cite{yin2018geonet} learned the 3D scene geometry from monocular videos to simultaneously estimate depth, pixel movement, and camera movement.

\subsection{Unsupervised / Self-Supervised Monocular Depth Estimation}\label{sub-unsupervise}
As ground truth is expensive to acquire in monocular depth estimation, many researchers focus on unsupervised approaches.
\cite{shu2020feature} proposed feature-metric loss to replace photometric loss and the authors used three different neural networks to estimate depth, image feature, and pose. 
\cite{watson2021temporal} proposed an adaptive approach that can make use of sequence information. They also proposed a consistency loss that can make the network ignore unreliable cost volume.
\cite{casser2019depth} proposed a framework for learning depth map, camera ego-motion, and object motions by representing objects in 3D, modeling its motion as SE3 transforms, and incorporating the SE3 transforms into the training process. 
\cite{bozorgtabar2019syndemo} proposed a view decoder to capture the 3D scene structure as a supervised signal and use that to jointly train a single-view depth estimation network and a pose estimation network.

\subsection{Monocular Depth Estimation in Dynamic Scenes}
Although all the methods in  \ref{sub-supervise} and \ref{sub-unsupervise} perform well, they do not consider real-world dynamic scenes.
Because dynamic scenes break static world assumption, many methods focus on how to recover depth value when both the camera and the scene are in motion. 
\cite{ranftl2016dense} proposed a two stage method for the same. The first stage performs motion segmentation and the second stage jointly reasons about the scales of different components and their location relative to the camera. This method is motivated by optical flow estimation and hence will fail if optical flow estimation fails.
\cite{di2019monocular} proposed a framework for monocular depth estimation in dynamic scenes by using super-pixel relations between neighboring frames in a monocular video. To solve the scale ambiguity problem in super-pixel relations, the authors assumed that motion and spatial relations between neighboring super-pixels correspond one-to-one.
\cite{li2020unsupervised} presented a method for training the estimation of depth, ego-motion, and a dense 3D translation field of objects in dynamic scenes.
The authors assumed that the dynamic scene is sparse, i.e., most of the scene is static, and they tend to be piece-wise constant for rigid moving objects.
However, as the authors pointed out, they did not consider the dense dynamic scenes.
Our proposed approach does not make such assumptions on sparseness of the dynamic scene.

\begin{figure*}[t]
	\centering
	\includegraphics[width=1\linewidth]{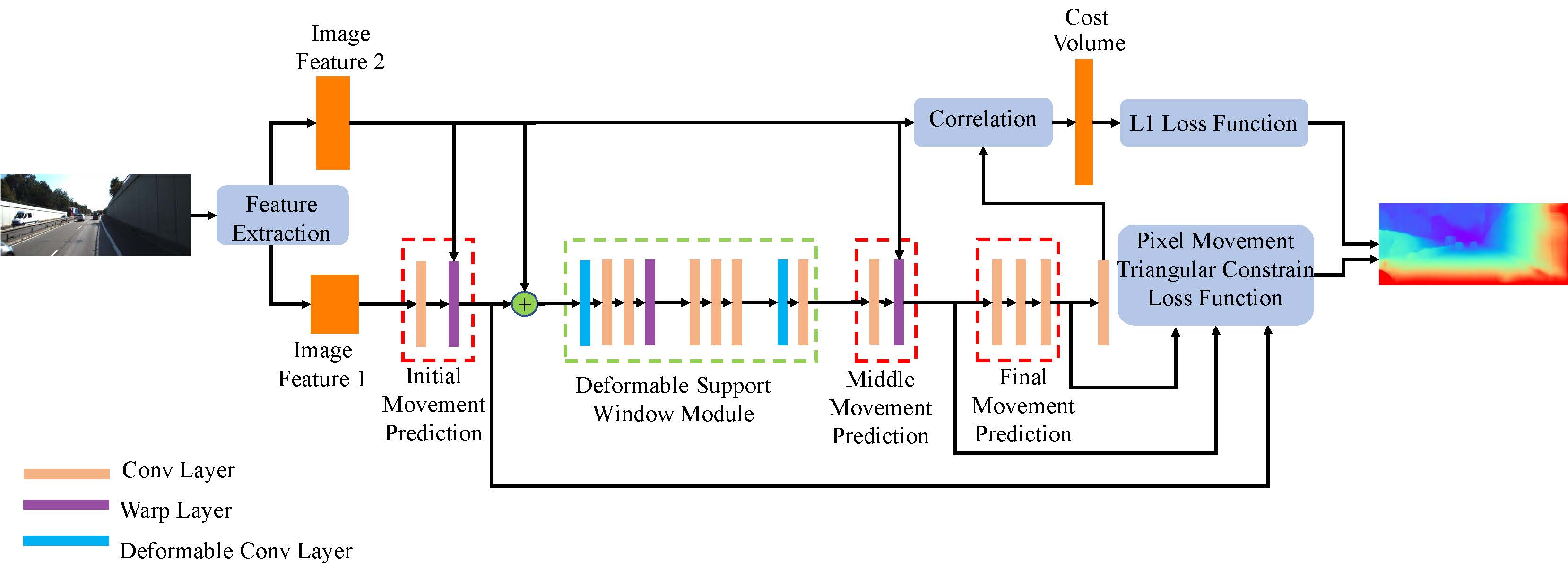}
	\vspace{-5mm}
	\caption{\textbf{Overall Network Architecture:} The model first extracts two features $F_1$ and $F_2$ at $1/6^{th}$ and $1/12^{th}$ scale respectively. These features are passed through the deformable support window module (green dotted box) and the 3-part pixel movement prediction module (red dotted boxes) to finally obtain the cost volume for depth prediction. The overall loss function consists of the traditional depth L1 loss and a Pixel Movement Triangular Constraint loss.}
	\vspace{-3mm}
	\label{network}
\end{figure*}

\section{PMPNet - Pixel Movement Prediction Network}\label{Method}

Our proposed neural network architecture PMPNet, as shown in Figure \ref{network}, consists of two main modules - individual pixel movement prediction and deformable support window.
Given an input RGB image $I$, we first extract the feature pyramid \{$F_1,F_2$\} at $1/6^{th}$ and $1/12^{th}$ resolutions respectively.
Next, we predict the initial pixel movement $PM_1$ by passing in the feature $F_2$ and the convolved feature $F_1$ to a warp layer.
This initial pixel movement $PM_1$ is concatenated in the channel direction with the feature $F_2$.
Next, a novel deformable support window approach is used to predict a second pixel movement $PM_2$ that better incorporates the compound transformations, especially at the edges.
We then perform multiple convolutions as part of the final pixel movement prediction $PM_3$.
This $PM_3$ is used to better constrain the first two pixel movements using a triangular loss function.
The final pixel movement $PM_3$ is convolved and passed together with the input image feature $F_2$ to construct the cost volume.
This cost volume is computed similar to \cite{xu2020aanet}, using the following equation:

\begin{equation}\label{eq:cost volume}
C(d,x,y) = \frac{1}{N} \left \langle PM_3(x,y),F_2(x,y-d) \right \rangle
\end{equation}    

During our 3-part pixel movement prediction, we up-sample the feature $F_2$ before concatenating with the feature $F_1$.
Because of this, the computed cost volume will inherently contain the features of the original left image at two different scales.
Hence, we do not need to perform a multi-scale aggregation step as done in \cite{xu2020aanet}, thereby saving us some training time.
The cost volume is then used to estimate the final depth as proposed within the deformable support window module.



\subsection{Pixel Movement Prediction Module} \label{ofp}
Previous works \cite{ranjan2019competitive, yin2018geonet} used the input image and additional information such as monocular video to predict the pixel movements.
These approaches assume that all objects and pixels in the scene are static.
Rather, we lift that assumption and predict different possible movements for each pixel.
However, without that assumption, it is hard to estimate depth value unless some constraints are added on the object/pixel trajectory. 
Similar to \cite{avidan2000trajectory}, we assume that the pixel moves along a straight-line and the camera center of projection traces a straight line during camera motion.
With these two assumptions, there exist two two-dimensional vectors, named $v_1$, $v_2$ which span a null space and a straight line $L$ could be represented in that null space by $L \cong \lambda_1 v_1 + \lambda_2 v_2$. 
As different CNN layers represent different scales, two parameters $\lambda_1$ and $\lambda_2$ are needed for the prediction of $v_1$ from $F_1$, $v_2$ from $F_2$, and $L$ from $v_1$ and $v_2$.

Figure \ref{network} shows all three pixel movement prediction parts. 
Based on where the pixel movements are incorporated in our model, we call them as initial or $v_1$, middle or $v_2$, and final or $L$.
The initial and middle pixel movement represent two possible move directions of a pixel. 
Such two possible move directions have good generalization because they could point to any direction.

Also we propose a Pixel Movement Triangular Constraint (PMTC) Loss according to $L \cong \lambda_1 v_1 + \lambda_2 v_2$ as defined by the following equation:
\begin{equation}\label{T}
	L_{PMTC} = \mathcal{L}(v_{1}+v_{2}, L) 
\end{equation}

The final loss function for our monocular depth estimation model is obtained by combining our proposed Pixel Movement Triangular Constraint Loss $L_{PMTC}$ with the standard L1 depth loss used in \cite{xu2020aanet}.

As shown in Figure \ref{network}, our initial and middle pixel movement prediction parts consist of two layers - convolution and warping.
The prediction of the initial pixel movement $v_1$ is performed on the features extracted from the input image.
The middle pixel movement $v_2$ is predicted on the output of the deformable support window module.
$PM_2$ is passed into the final pixel movement prediction part, which consists of three traditional convolution layers.
This final pixel movement $L$ is convolved again and passed to the correlation step alongside the input image feature $F_2$ to compute the cost volume for the depth L1 loss function.

\subsection{Deformable Support Window Module}\label{DSW} 
Majority of the monocular depth estimation method suffers from the depth inconsistency problem of blurring the edge regions.
To overcome this, we use a deformable support window module, as shown in Figure \ref{network}.
This idea is motivated by the property of deformable convolutions that it enables free form deformation of the sampling grid, which would lead to better learning of compound transformations \cite{dai2017deformable}.
The use of a deformable support window is inspired and built on top of \cite{bleyer2011patchmatch} who proposed a slanted support window to compute disparity more accurately for a slanted surface such as a corridor, and \cite{tankovich2020hitnet} who used a CNNs layer to learn the slanted support window.
However, our support window module is different from \cite{tankovich2020hitnet} in that our method uses the deformable convolution layer to learn depth, which performs better near edge areas.
The use of these deformable convolution layers provides better depth estimates near the edges, thereby addressing the problem of having blurry-edges.

\begin{table*}[b]
\vspace{2mm}
\small
\begin{center}
\setlength{\tabcolsep}{4mm}
\renewcommand{\arraystretch}{1.3} 
\begin{tabular}{|l|c|c|c|c||c|c|c|}
\hline
\multicolumn{1}{|c|}{\multirow{2}{*}{Method}}     & Abs Rel        & Sq Rel         & RMSE           & RMSE $log$     & $\delta < 1.25$      & $\delta < 1.25^{2}$ & $\delta < 1.25^{3}$ \\ \cline{2-8} 
\multicolumn{1}{|c|}{}                                  & \multicolumn{4}{c||}{Lower is better}                              & \multicolumn{3}{c|}{Higher is better}                            \\ \hline \hline
Eigen\cite{eigen2014depth}             & 0.203          & 1.548          & 6.307          & 0.282          & 0.702                & 0.890               & 0.890               \\ \hline
Yin$*$ \cite{yin2019enforcing}            & 0.155          & 1.296          & 5.857          & 0.233          & 0.793                & 0.931               & 0.973               \\ \hline
Garg$*$ \cite{garg2016unsupervised}       & 0.152          & 1.226          & 5.849          & 0.246          & 0.784                & 0.921               & 0.967               \\ \hline
DDVO$*$ \cite{wang2018learning}           & 0.151          & 1.257          & 5.583          & 0.228          & 0.810                & 0.936               & 0.974               \\ \hline
EPC++$*$ \cite{luo2019every}              & 0.141          & 1.029          & 5.350          & 0.216          & 0.816                & 0.941               & 0.976               \\ \hline
mono2$*$ \cite{godard2019digging}         &  0.115         & 0.903          & 4.863          & 0.193          & 0.877                & 0.959               & 0.981               \\ \hline
DORN \cite{fu2018deep}                 & 0.072          & 0.307          & 2.727          & 0.120          & 0.932                & 0.984               & 0.994               \\ \hline
FeatureDepth$*$\cite{shu2020feature}      &0.079             &0.666 
&3.922            &0.163          &0.925      &0.970 &0.984\\ \hline
SC-GAN\cite{wu2019spatial}             & 0.063    &  0.178   &  2.129    & 0.097 &  0.961          &  0.993       &  0.998              \\ \hline
DPT-Hybrid\cite{ranftl2021vision}             &  0.062    &  0.198    &  2.573   & 0.092 & 0.959          &  0.995        &  0.999               \\ \hline
AdaBins\cite{bhat2021adabins}             &  0.058    & 0.190    &  2.360    &0.088 &  0.964          &  0.995         &  0.999               \\ \hline
SingleNet$*$\cite{chen2021revealing}             &  0.094    &0.681     &4.392     &0.185  &0.892           &0.965          & 0.981              \\ \hline
You\cite{you2021towards}             & 0.058     &  0.194   &2.415     & 0.092 &  0.960          & {\ul 0.995}         &  0.999             \\ \hline
ManyDepth\cite{watson2021temporal}             &0.087     &0.685     &4.142    &0.167  &0.920            &0.968          &0.983              \\ \hline
EdgeConv\cite{lee2022edgeconv}             &{\ul0.056}     & {\ul 0.169}     &{\ul 1.925}     &{\ul 0.087}  &{\ul 0.964}            &0.994          &{\ul 0.999}              \\ \hline
Ours                                      & \textbf{0.049} & \textbf{0.157} & \textbf{1.831} &\textbf{0.071}  & \textbf{0.994}       & \textbf{0.996}      &\textbf{0.999}      \\ \hline
\end{tabular}
\end{center}
    \vspace{-2mm}
	\caption{\textbf{Quantitative Results - KITTI \cite{geiger2012we}:} Comparison of our model on the KITTI test split by Eigen \cite{eigen2014depth}. Best results for each metric are in \textbf{bold}; second best are \uline{underlined}. All results presented here are without post-processing. The $\ast$ denotes that the approach is either self-supervised or unsupervised, and the remaining are supervised approaches.}
 	\vspace{-3mm}
\label{KITTI quan}
\end{table*}

\subsection{Depth Regression}\label{Sub-pixel Disparity}
To estimate the depth image, we propose a new depth generation approach.
Compared with \cite{bhat2021adabins, xu2020aanet}, our formula combines the offsets from the deformable convolution layer, which learn the compound geometric transformations near the edges.
We first apply softmax \cite{kendall2017end} to build an initial depth $\widetilde{d}(p_{i,j})$ for each pixel $p_{i,j}$.
This depth value can be given by the following equation:

\vspace{-3mm}
\begin{equation}
	\widetilde{d}(p_{i,j}) = \sum_{d=0}^{ N-1 } d(p_{i,j}) \times \sigma(C(d(p_{i,j}))) 
\end{equation}

where $N$ is the number of channels for the cost volume layer.
$\sigma$ is the softmax function and $C(d(p_{i,j}))$ is the matching cost for $p_{i,j}$ with depth $d$.

Next, we use this initial depth $\widetilde{d}(p_{i,j})$ in our new depth equation as follows:

\vspace{-5mm}
\begin{equation}\label{final d}
	d_{i,j} = (1 - \alpha) \widetilde{d}(p_{i,j}) + \alpha \sum_{k=1}^{8}d( p_{m,n} + \Delta_k(i,j))
\end{equation}

where $ \Delta_k(i,j)$ is the k-th offset of $p_{i,j}$ from the deformable convolution layer used in the proposed Deformable Support Window Module.
As we select the support window of size $3 \times 3$, hence there can only be $8$ offsets.
$d( p_{m,n} + \Delta_k(m,n))$ represents the depth of one of the 8 neighboring pixels of $p_{i,j}$ plus an offset $\Delta$.
The final depth values for all pixels are regressed by using a traditional L1 loss and the proposed Pixel Movement Triangular Constraint Loss $L_{PMTC}$, represented by Equation \ref{T}.

\begin{figure*}[t]
\vspace{3mm}
	\begin{center}
	    \subfigure{
		    \parbox[c]{2mm}{\rotatebox[origin=c]{90}{\small Input}}
		}
		\subfigure{
			\parbox[c]{0.23\linewidth}{\includegraphics[width=11.5em]{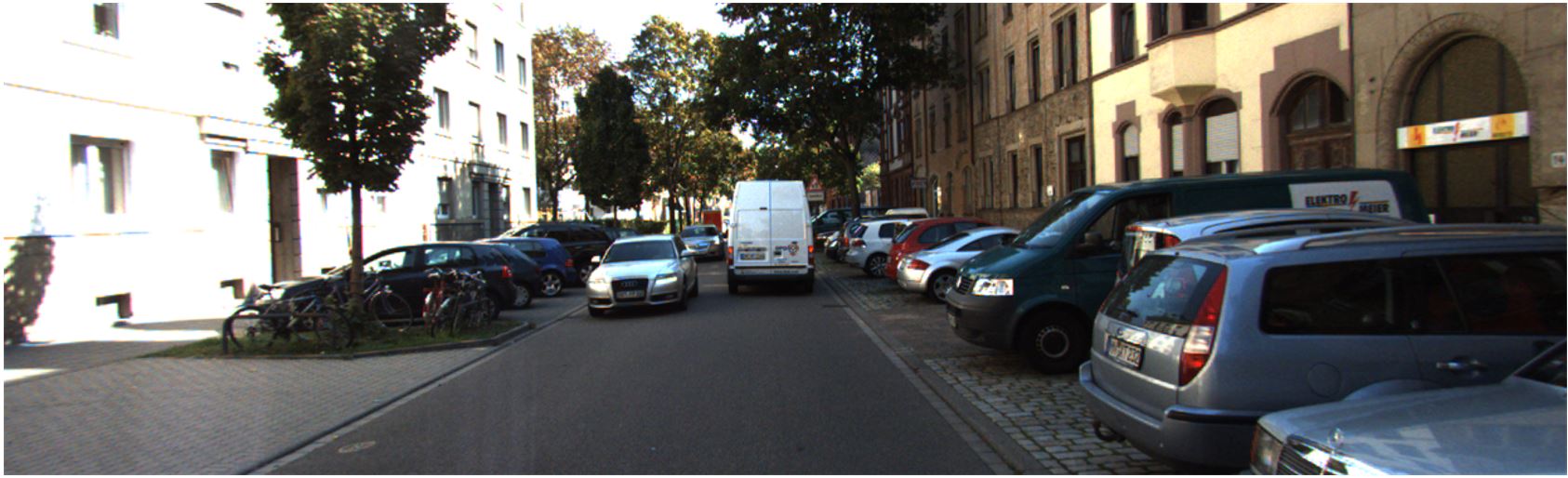}}
		}\hspace{-0.22in}
		\quad
 		\subfigure{
 			\parbox[c]{0.23\linewidth}{\includegraphics[width=11.5em]{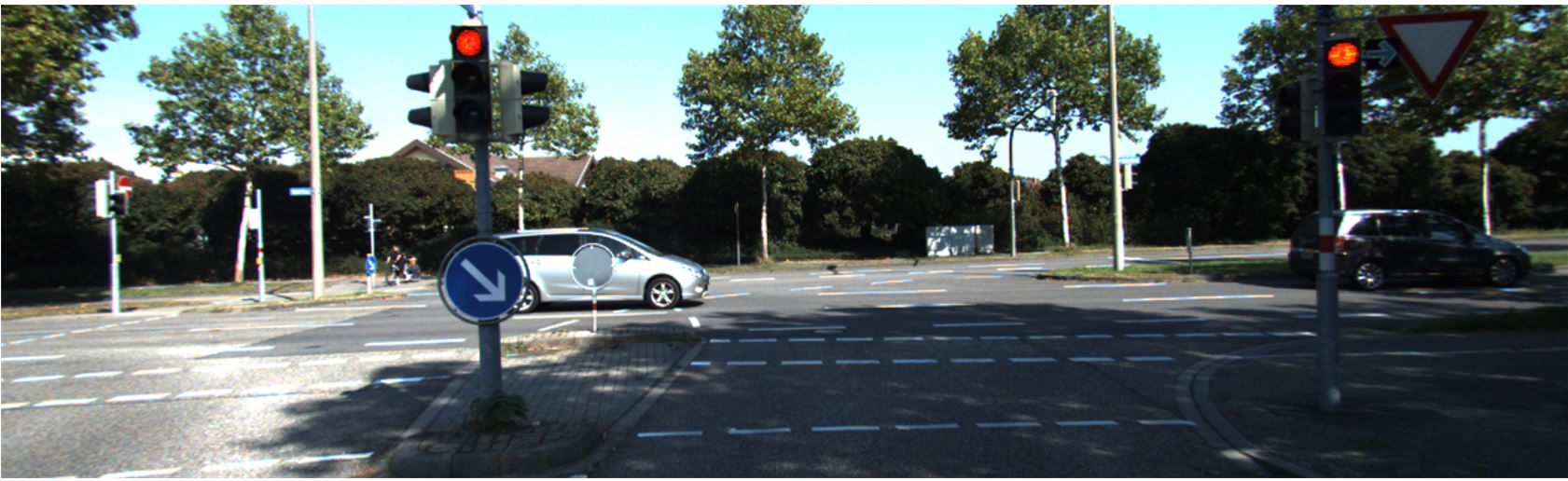}}
 		}\hspace{-0.22in} 
 		\quad
 		\subfigure{
 			\parbox[c]{0.23\linewidth}{\includegraphics[width=11.5em]{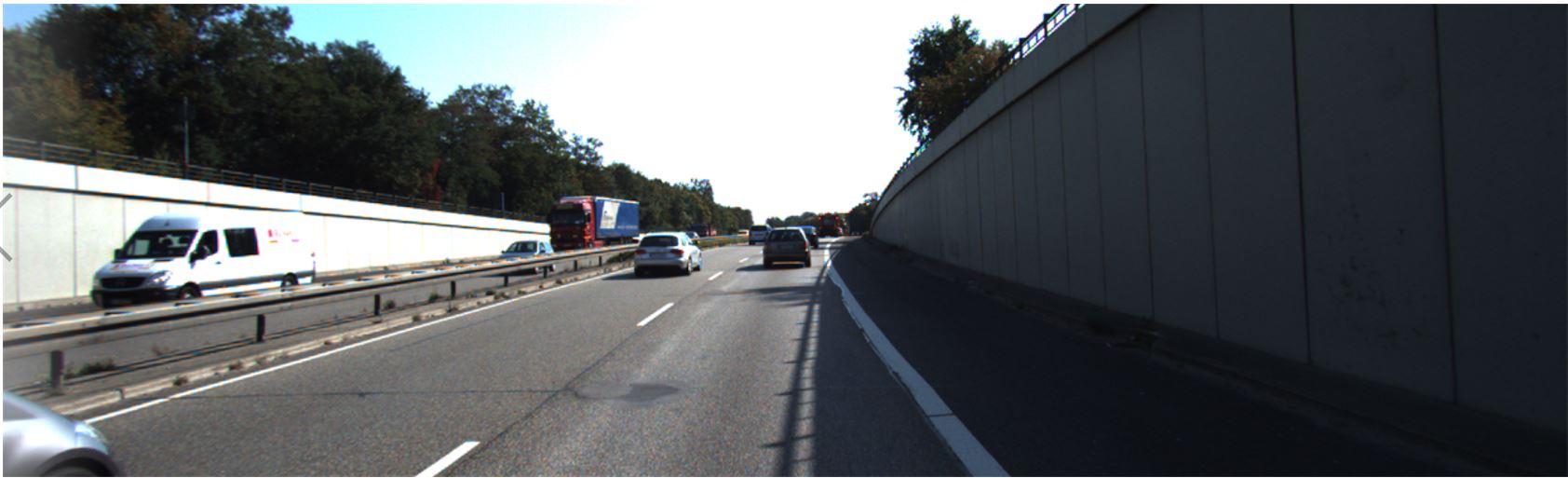}}
 		}\hspace{-0.22in} 
 		\quad
 		\subfigure{ 
 			\parbox[c]{0.23\linewidth}{\includegraphics[width=11.5em]{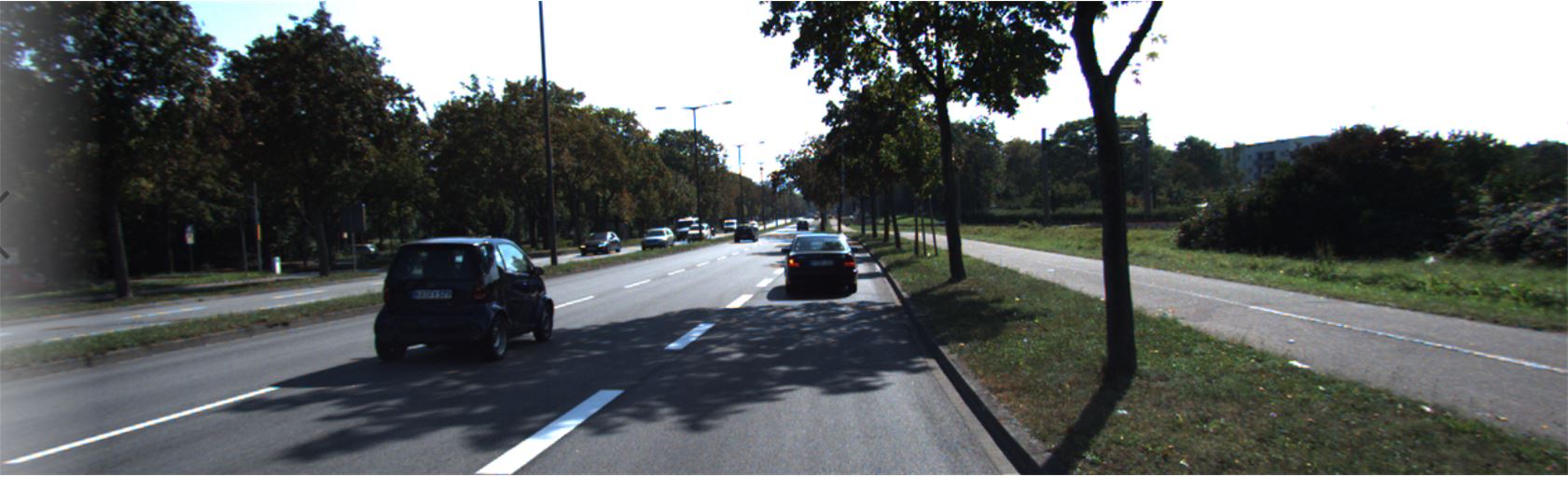}}
 		}\hspace{-0.22in} 
		\quad
		\vspace{-0.1in}

		\subfigure{
		    \parbox[c]{2mm}{\rotatebox[origin=c]{90}{\footnotesize Zhou\cite{zhou2017unsupervised}}}
		}
		\subfigure{
			\parbox[c]{0.23\linewidth}{\includegraphics[width=11.5em]{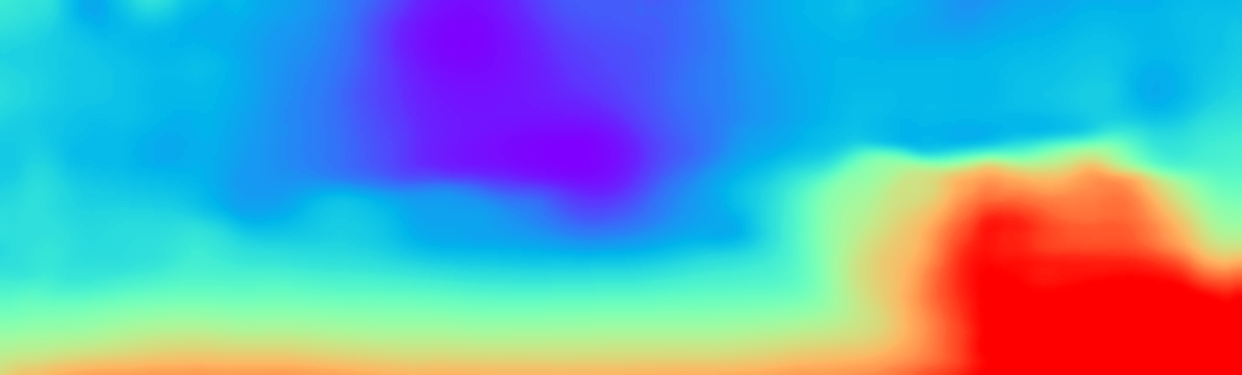}}
		}\hspace{-0.22in}
		\quad
		\subfigure{
			\parbox[c]{0.23\linewidth}{\includegraphics[width=11.5em]{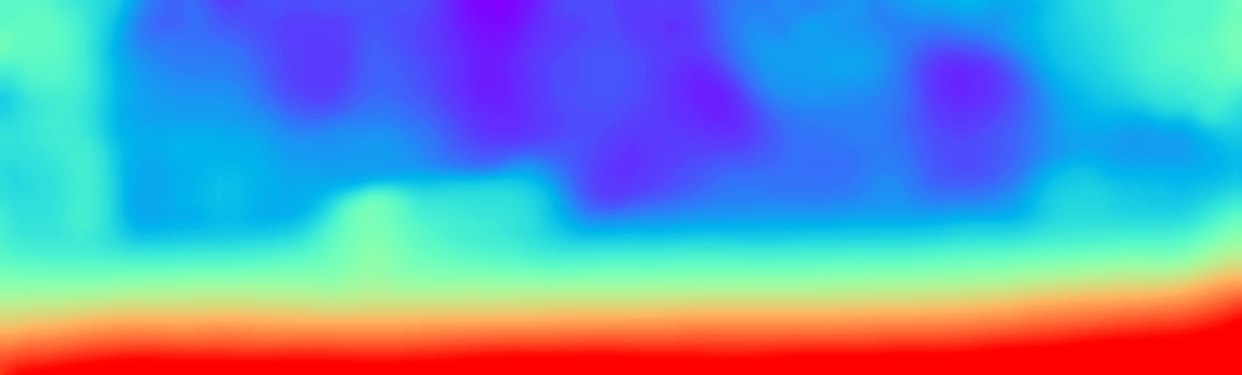}}
		}\hspace{-0.22in}
		\quad
		\subfigure{
			\parbox[c]{0.23\linewidth}{\includegraphics[width=11.5em]{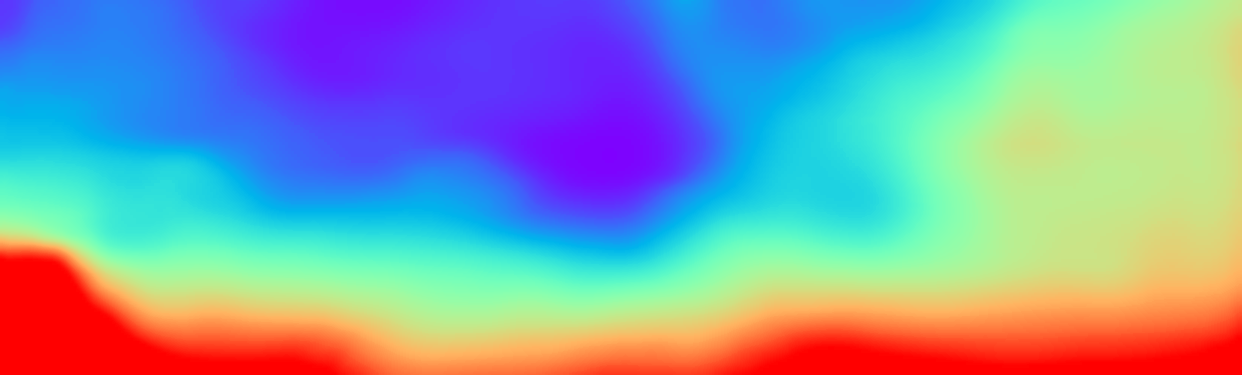}}
		}\hspace{-0.22in}
		\quad
		\subfigure{
			\parbox[c]{0.23\linewidth}{\includegraphics[width=11.5em]{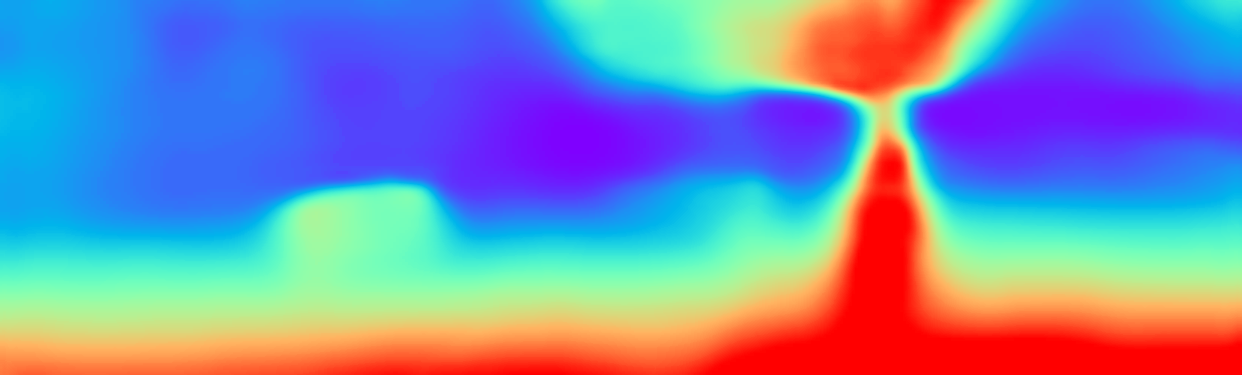}}
		}\hspace{-0.22in}
		\quad
		\vspace{-0.1in}

 		\subfigure{
 		    \parbox[c]{2mm}{\rotatebox[origin=c]{90}{\footnotesize DDVO\cite{wang2018learning}}}
		}
		\subfigure{
 			\parbox[c]{0.23\linewidth}{\includegraphics[width=11.5em]{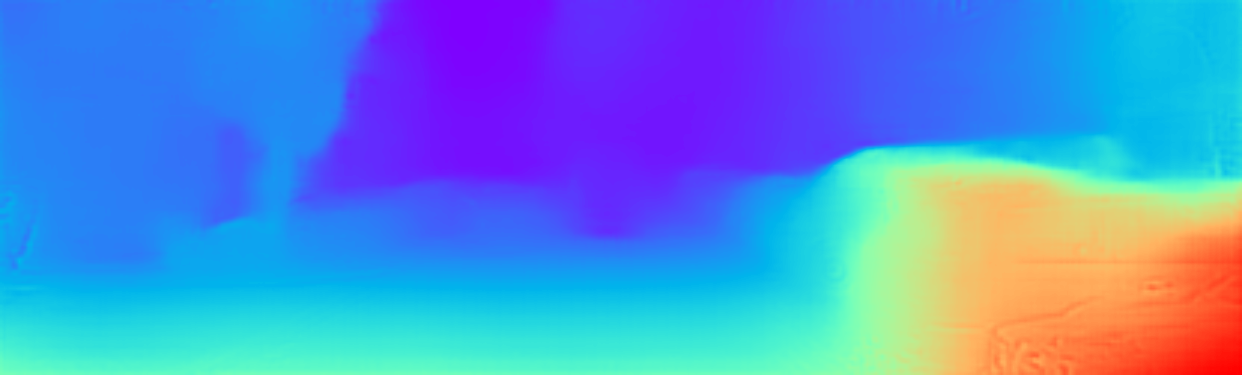}}
 		}\hspace{-0.22in}
 		\quad
		\subfigure{
			\parbox[c]{0.23\linewidth}{\includegraphics[width=11.5em]{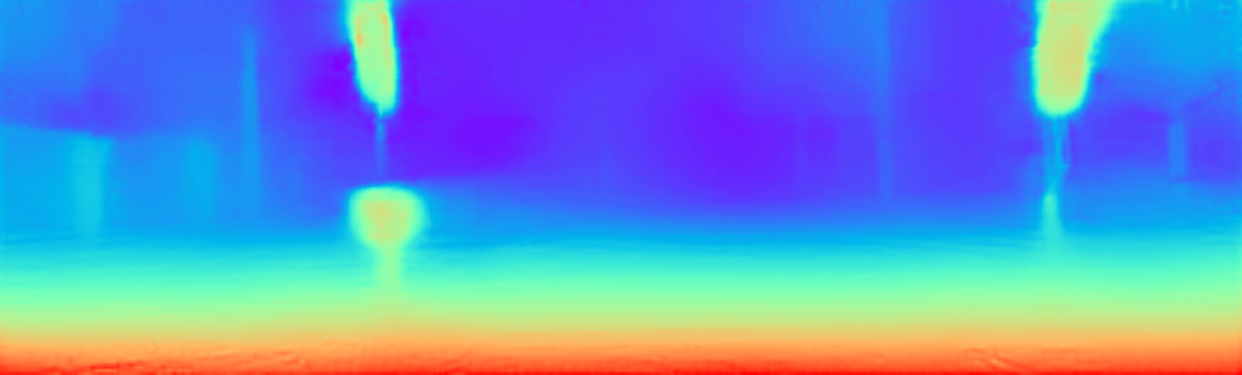}}
		}\hspace{-0.22in} 
		\quad
		\subfigure{
			\parbox[c]{0.23\linewidth}{\includegraphics[width=11.5em]{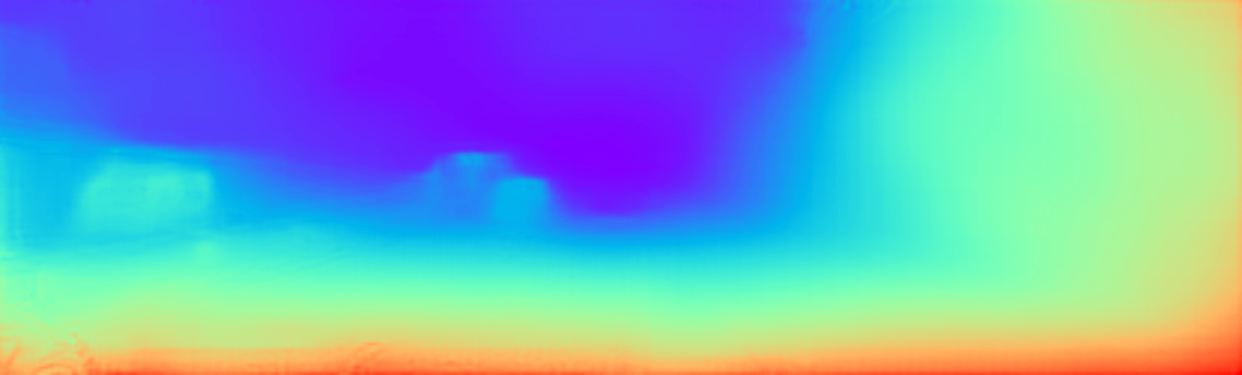}}
		}\hspace{-0.22in} 
		\quad
		\subfigure{
			\parbox[c]{0.23\linewidth}{\includegraphics[width=11.5em]{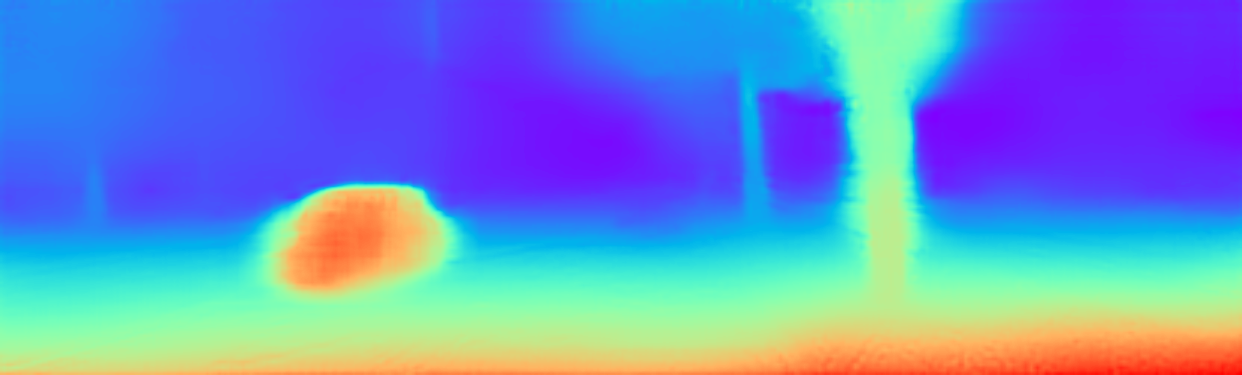}}
		}\hspace{-0.22in}
		\quad
		\vspace{-0.1in}

		\subfigure{
		    \parbox[c]{2mm}{\rotatebox[origin=c]{90}{\footnotesize mono\cite{godard2017unsupervised}}}
		}
		\subfigure{
			\parbox[c]{0.23\linewidth}{\includegraphics[width=11.5em]{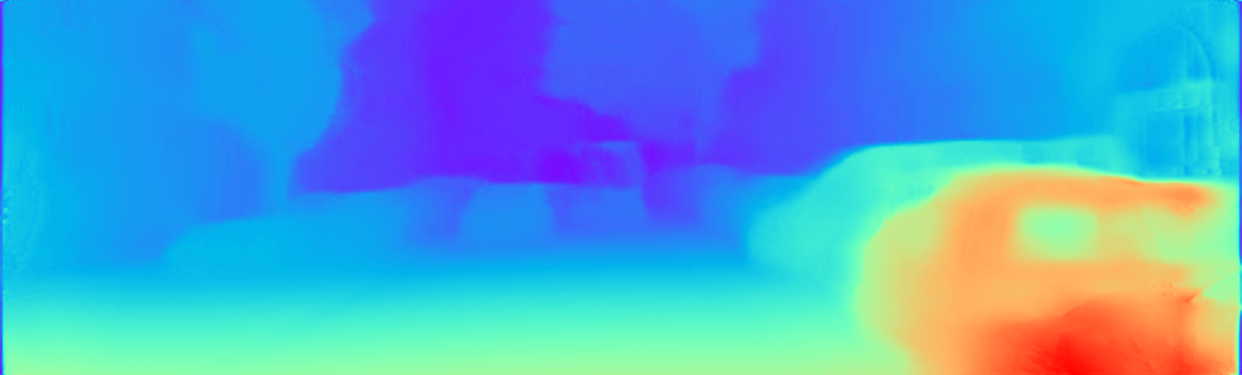}}
		}\hspace{-0.22in}
		\quad
		\subfigure{
			\parbox[c]{0.23\linewidth}{\includegraphics[width=11.5em]{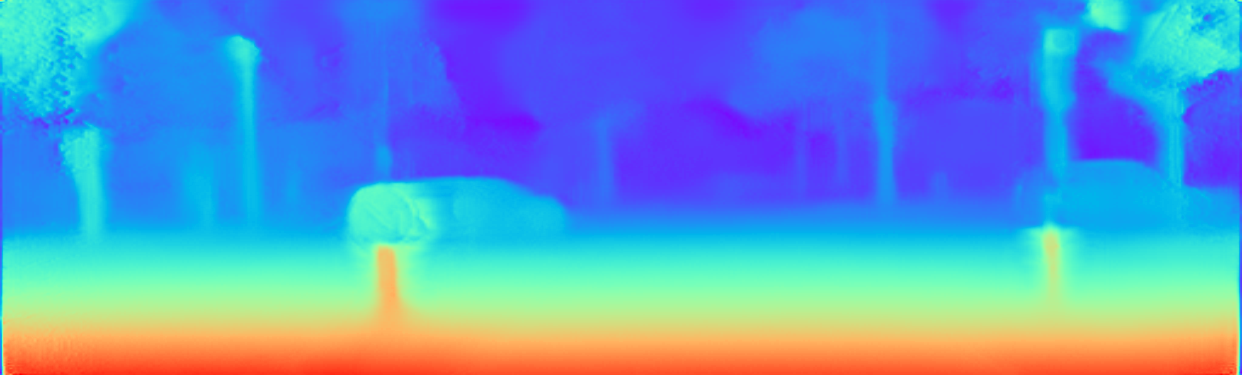}}
		}\hspace{-0.22in} 
		\quad
		\subfigure{
			\parbox[c]{0.23\linewidth}{\includegraphics[width=11.5em]{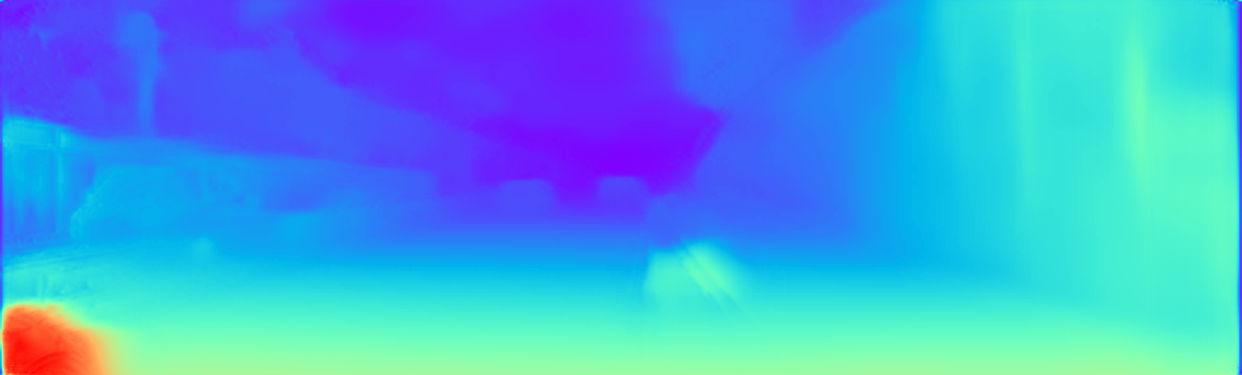}}
		}\hspace{-0.22in} 
		\quad
		\subfigure{
			\parbox[c]{0.23\linewidth}{\includegraphics[width=11.5em]{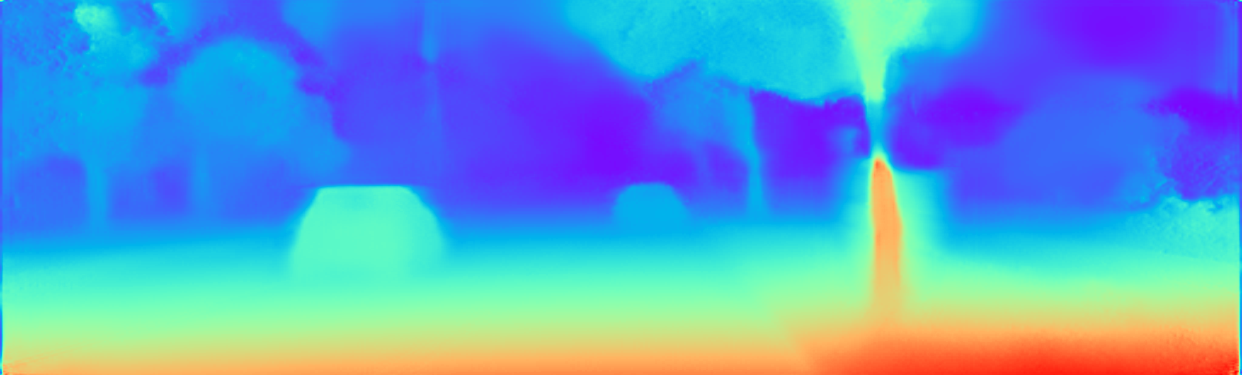}}
		}\hspace{-0.22in} 
		\quad
		\vspace{-0.1in}

		\subfigure{
		    \parbox[c]{2mm}{\rotatebox[origin=c]{90}{\footnotesize mono2\cite{godard2019digging}}}
		}
		\subfigure{
			\parbox[c]{0.23\linewidth}{\includegraphics[width=11.5em]{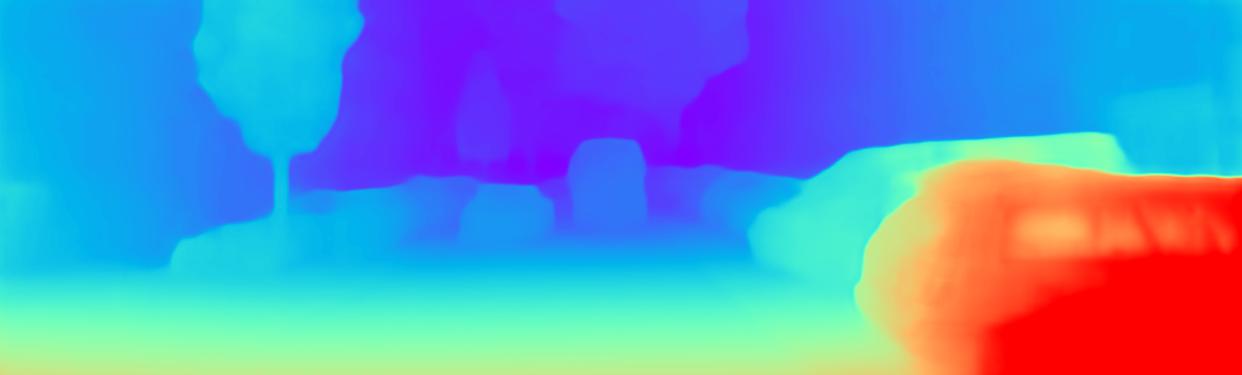}}
		}\hspace{-0.22in}
		\quad
		\subfigure{
			\parbox[c]{0.23\linewidth}{\includegraphics[width=11.5em]{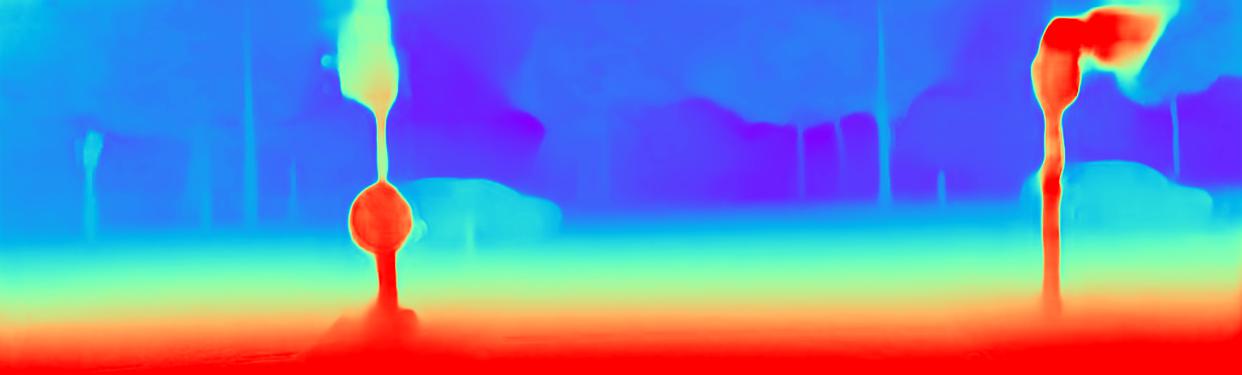}}
		}\hspace{-0.22in} 
		\quad
		\subfigure{
			\parbox[c]{0.23\linewidth}{\includegraphics[width=11.5em]{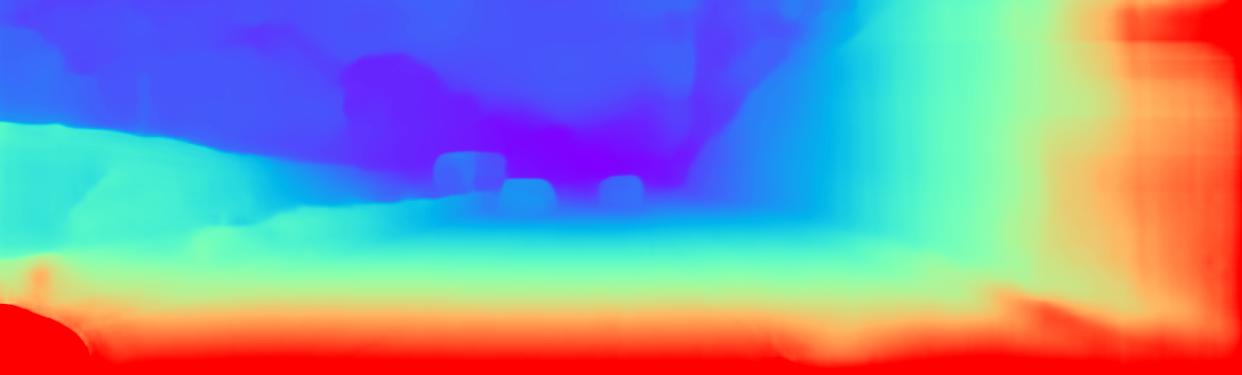}}
		}\hspace{-0.22in} 
		\quad
		\subfigure{
			\parbox[c]{0.23\linewidth}{\includegraphics[width=11.5em]{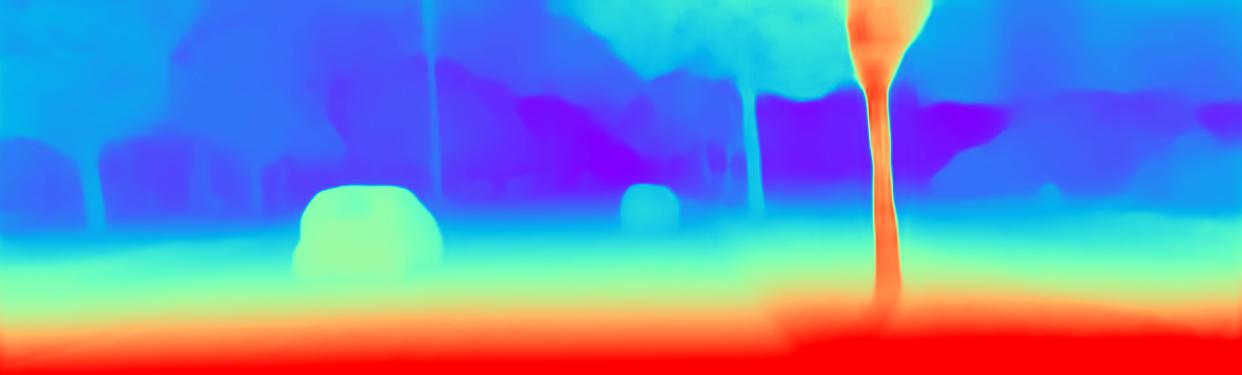}}
		}\hspace{-0.22in}
		\quad
		\vspace{-0.1in}

 		\subfigure{
 		    \parbox[c]{2mm}{\rotatebox[origin=c]{90}{\small Ours}}
		}
		\subfigure{
 			\parbox[c]{0.23\linewidth}{\includegraphics[width=11.5em]{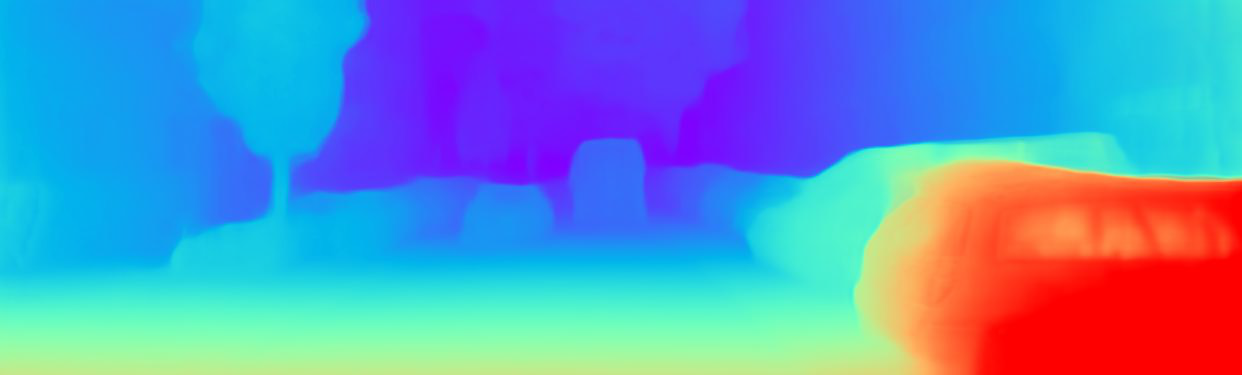}}
 		}\hspace{-0.22in}
 		\quad
 		\subfigure{
 			\parbox[c]{0.23\linewidth}{\includegraphics[width=11.5em]{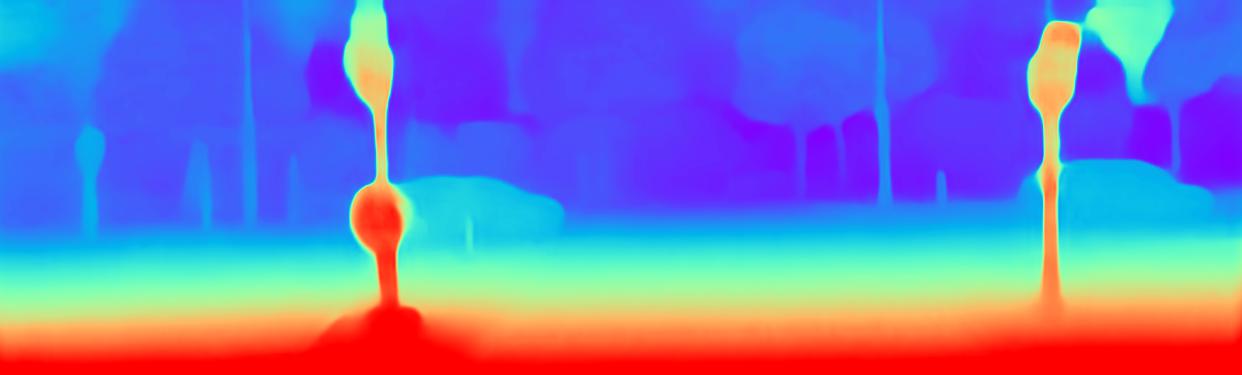}}
 		}\hspace{-0.22in}
 		\quad
 		\subfigure{
			\parbox[c]{0.23\linewidth}{\includegraphics[width=11.5em]{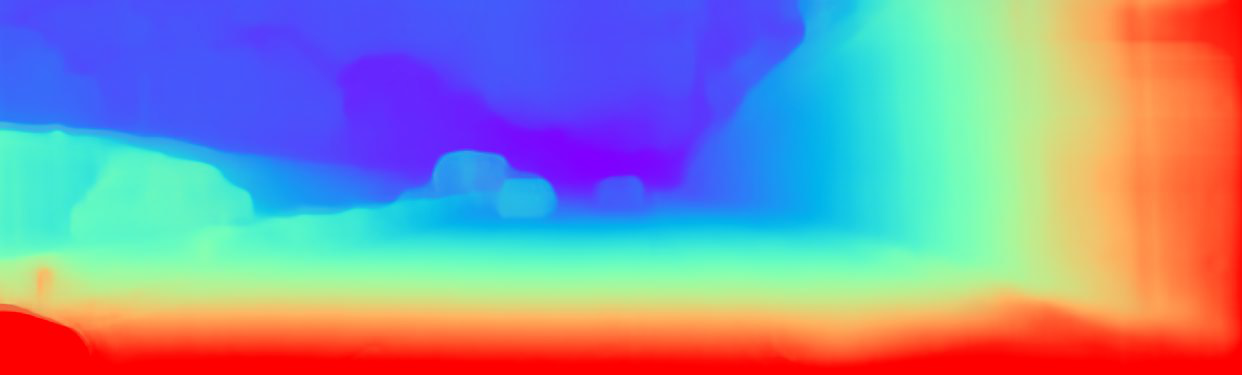}}
 		}\hspace{-0.22in}
 		\quad
 		\subfigure{
 			\parbox[c]{0.23\linewidth}{\includegraphics[width=11.5em]{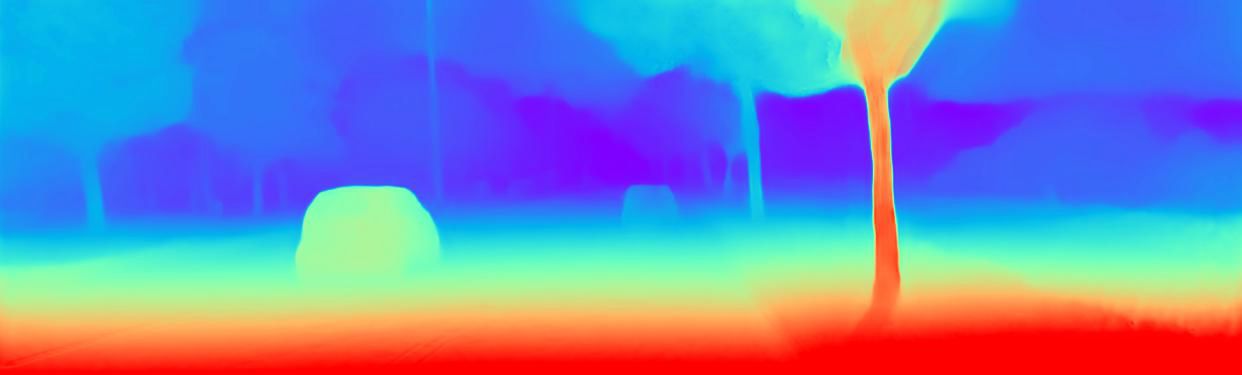}}
 		}\hspace{-0.22in} 
 		\quad
 		\vspace{-0.1in}
		\caption{\textbf{Qualitative Results - KITTI \cite{geiger2012we}:} Depth estimated on four different test images (first row) using our approach (last row) in comparison with four other methods (remaining rows).}
		\vspace{-5mm}
	\label{KITTI visual}
    \end{center}
\end{figure*}

\section{Experiments \& Results}\label{exp}
In this section, we evaluate our proposed model by conducting experimental analysis on two outdoor datasets: KITTI \cite{geiger2012we} and Make3D \cite{saxena2008make3d} as well as an indoor dataset: NYU Depth V2 \cite{silberman2012indoor}.
For each dataset we report two types of results, quantitative and qualitative, in comparison with other state-of-the-art monocular depth estimation methods.

For \textit{quantitative} analysis, the evaluation metrics used are as in the previous works \cite{eigen2014depth,godard2019digging}.
For the KITTI dataset, we use seven metrics - Abs Rel, Sq Rel, RMSE, RMSE $log$, $\delta < 1.25$, $\delta < 1.25^{2}$, and $\delta < 1.25^{3}$.
For the Make3D dataset and NYU Depth V2, we use - Abs Rel, Sq Rel, RMSE, and RMSE $log$.

For \textit{qualitative} analysis, we compare the visual results of our approach for a sample image against the other methods.

\subsection{Implementation Details}
Our model is built on the PyTorch framework \cite{paszke2019pytorch}.
We train our model on the KITTI dataset using a single NVIDIA RTX 3080 GPU for 60 epochs with a batch size of 4.
Adam optimizer \cite{kingma2014adam} is used with the settings $\beta_{1} = 0.9$ and $\beta_{2} = 0.999$.
We use an adaptive learning rate, starting at $0.001$ and decreasing it by half every 10 epochs after the 20th epoch.

\subsubsection{KITTI Dataset Details \cite{geiger2012we}} \label{kitti-dataset}
From the KITTI dataset \cite{geiger2012we}, we use all the 61 scenes from the “City”, “Residential”, “Road”, and “Campus” categories in the raw data as our training/testing sets.
We use the split by Eigen et al. \cite{eigen2014depth}, where we test our model on 697 images from the 29 scenes and train on 10K images picked uniformly from the remaining 32 scenes.
For efficient computation, all input images are resized to $336 \times 960$.
We cap the depth to $80m$ during evaluation.

\subsubsection{Make3D Dataset Details \cite{saxena2008make3d}}
We use our model pre-trained on the KITTI dataset and fine-tune the parameters before testing it on all 534 images in the Make3D dataset.
As the size of the ground truth $55 \times 305$ and the size of the input image $1704 \times 2272$ are not aligned, we center crop the input image with a ratio of $1 \times 2$ and then resize it to $336 \times 960$.
The ground truth image is also center cropped with the same ratio.

\begin{table}[]
\vspace{2mm}
\small
\setlength{\tabcolsep}{2.3mm}
\renewcommand{\arraystretch}{1.3} 
\vspace{-2mm}
\begin{tabular}{|l||c|c|c|c|}
\hline
\multicolumn{1}{|c||}{\multirow{2}{*}{Method}}      & Abs Rel        & Sq Rel         & RMSE           & RMSE $log$     \\ \cline{2-5} 
\multicolumn{1}{|c||}{}   & \multicolumn{4}{c|}{Lower is better}   \\ \hline \hline
mono* \cite{godard2017unsupervised}    & 0.443            & 7.112           & 8.860           & 0.142            \\ \hline
DDVO$*$ \cite{wang2018learning}        & 0.387            & 4.720           & 8.090           & 0.204            \\ \hline
mono2$*$ \cite{godard2019digging}      & 0.322            & 3.589           & 7.417           & 0.163            \\ \hline
SC-GAN\cite{wu2019spatial}             & 0.206            & 3.550           & 5.478           & 0.801      \\ \hline
DORN \cite{fu2018deep}            &0.197             & 3.461           &5.671            & 0.746      \\ \hline
DPT-Hybrid\cite{ranftl2021vision}             & 0.182            & 3.021           &5.773            &0.766       \\ \hline
AdaBins\cite{bhat2021adabins}              &0.158             &2.874            & 5.355           & 0.702      \\ \hline
SingleNet$*$\cite{chen2021revealing}               & 0.146            & 2.287           & 5.536           & 0.728      \\ \hline
You\cite{you2021towards}                &0.120             &2.258            &6.471            &0.248      \\ \hline
ManyDepth$*$\cite{watson2021temporal}     &0.119  &1.857  & 7.633   & 0.195      \\ \hline

EdgeConv\cite{lee2022edgeconv}         & 0.125   &1.987   &5.105   &0.077  \\ \hline
Ours                                   & \textbf{0.101}   & \textbf{1.736}  & \textbf{ 4.337}  & \textbf{0.062}   \\ \hline
\end{tabular}
	\caption{\textbf{Quantitative Results - Make3D \cite{saxena2008make3d}:}Best results for each metric are in \textbf{bold}; second best are \uline{underlined}.}
	\label{make3d qua}
	\vspace{-4mm}
\end{table}

\begin{table}[]
\vspace{2mm}
\small
\setlength{\tabcolsep}{2.3mm}
\renewcommand{\arraystretch}{1.3} 
\vspace{-2mm}
\begin{tabular}{|l||c|c|c|c|}
\hline
\multicolumn{1}{|c||}{\multirow{2}{*}{Method}}      & Abs Rel        & Sq Rel         & RMSE           & RMSE $log$     \\ \cline{2-5} 
\multicolumn{1}{|c||}{}   & \multicolumn{4}{c|}{Lower is better}   \\ \hline \hline
mono* \cite{godard2017unsupervised}    & 0.476            &7.155            &9.368            &0.142             \\ \hline
DDVO$*$ \cite{wang2018learning}        &0.405             &5.271            &8.815            &0.413             \\ \hline
mono2$*$ \cite{godard2019digging}      & 0.352            &3.905            &7.141            &0.196           \\ \hline
SC-GAN\cite{wu2019spatial}             &0.401            &1.875            &5.367           &0.251       \\ \hline
DORN \cite{fu2018deep}            &0.374             &3.792           &8.238             &0.201      \\ \hline
DPT-Hybrid\cite{ranftl2021vision}             &0.110             & 0.335           & 0.357            & 0.045      \\ \hline
AdaBins\cite{bhat2021adabins}              &0.103             & 0.312           &0.364            &0.044       \\ \hline
SingleNet$*$\cite{chen2021revealing}               &0.138             &0.382            &  0.475         & 0.058      \\ \hline
You\cite{you2021towards}                &0.110             &0.301            &0.392            &0.047       \\ \hline
ManyDepth\cite{watson2021temporal}     &0.117             &1.182            & 6.031           & 0.179      \\ \hline

EdgeConv\cite{lee2022edgeconv}         & 0.107    & 0.298  & 0.373  & 0.046 \\ \hline
Ours                                   & \textbf{0.075}   & \textbf{0.105}  & \textbf{0.194}  & \textbf{0.028}   \\ \hline
\end{tabular}
	\caption{\textbf{Quantitative Results - NYU Depth V2 \cite{silberman2012indoor}:}Best results for each metric are in \textbf{bold}; second best are \uline{underlined}.}
	\label{NYU qua}
	\vspace{-4mm}
\end{table}

\subsubsection{NYU Depth V2 \cite{silberman2012indoor}}.  This is a dataset that provides images and depth maps for different indoor scenes captured at a pixel resolution of $640 \times 480$. The dataset contains 120K training samples and 654 testing samples. We train our
network on a 10K subset. The upper bound of depth maps is 10 meters. The resolution of our output is $320 \times 240 $. Then we upsample it to $640 \times 480$ to match the ground truth resolution during both
training and testing.

\begin{figure}[t]
	\begin{center}

		\subfigure{
		    \parbox[c]{2mm}{\rotatebox[origin=c]{90}{Input}}
		}
		\subfigure{
			\parbox[c]{0.1\linewidth}{\includegraphics[width=5em]{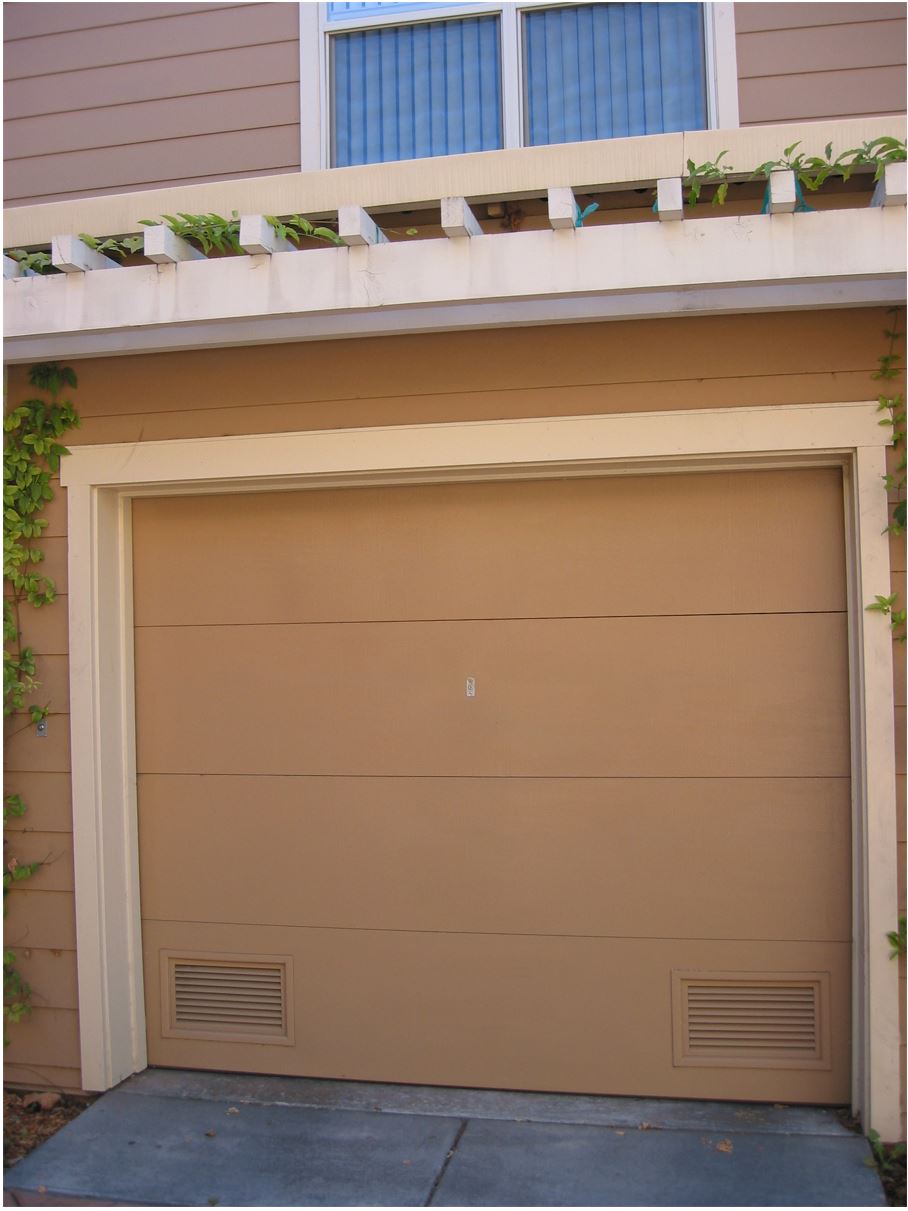}}
		}\hspace{0.1in}
		\quad
 		\subfigure{
 			\parbox[c]{0.1\linewidth}{\includegraphics[width=5em]{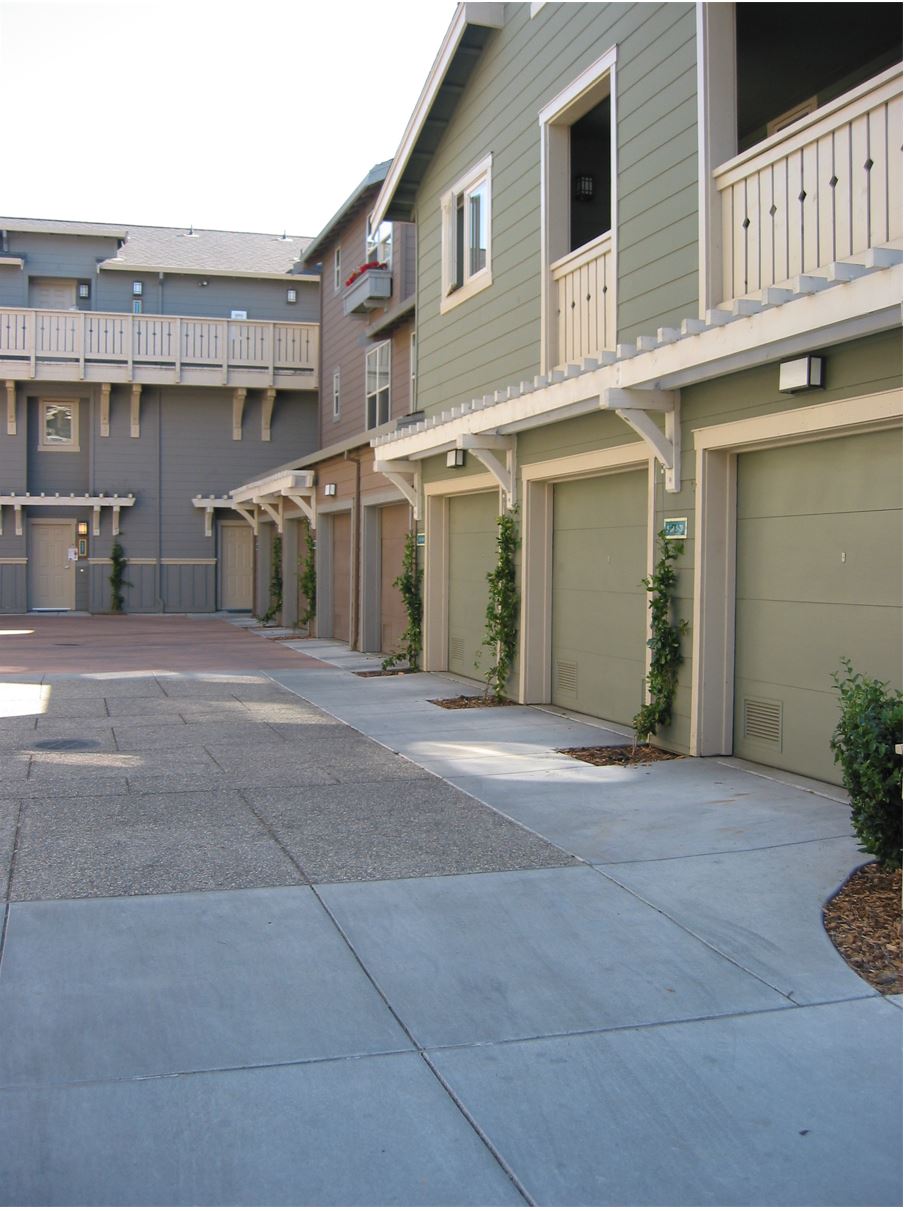}}
 		}\hspace{0.1in} 
 		\quad
 		\subfigure{
 			\parbox[c]{0.1\linewidth}{\includegraphics[width=5em]{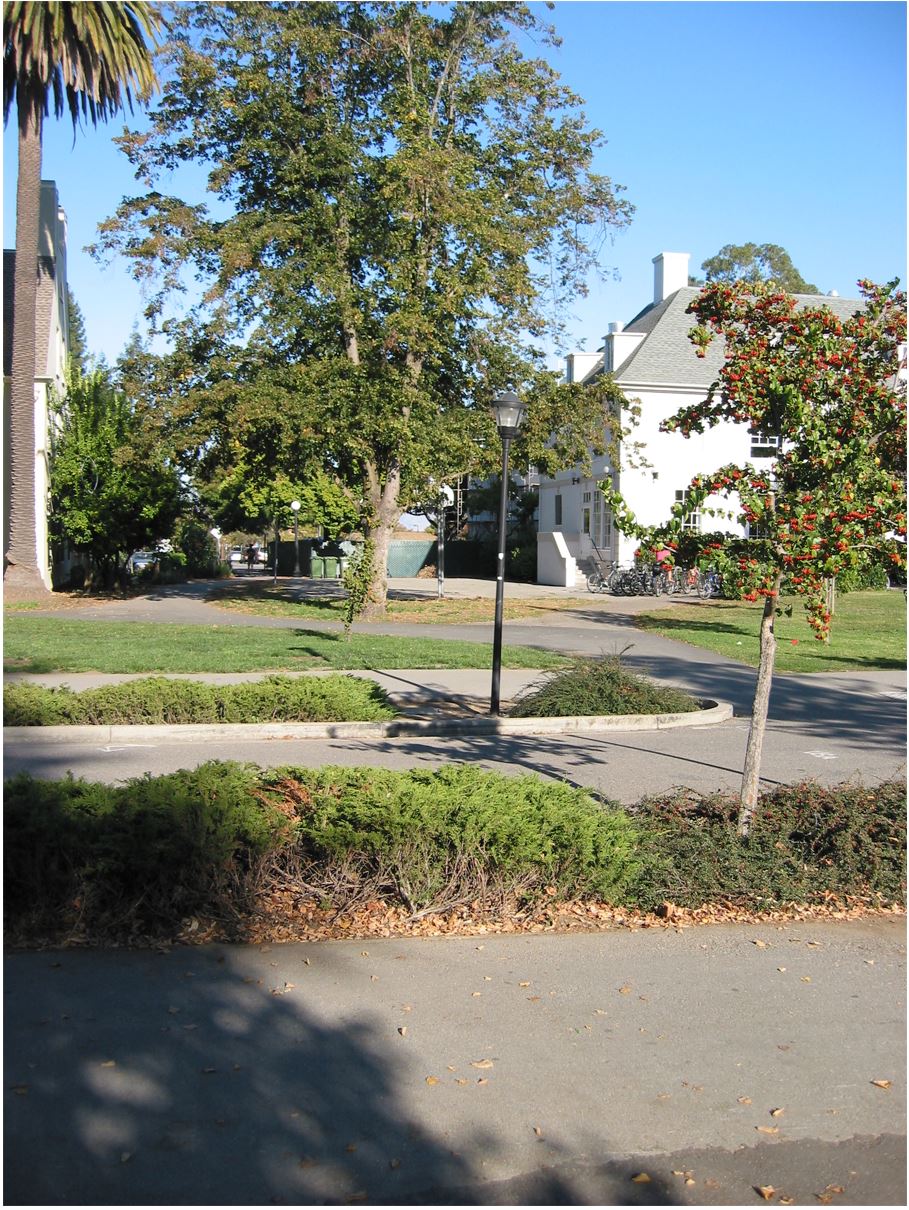}}
 		}\hspace{0.1in} 
		\quad
		\subfigure{
 			\parbox[c]{0.1\linewidth}{\includegraphics[width=5em]{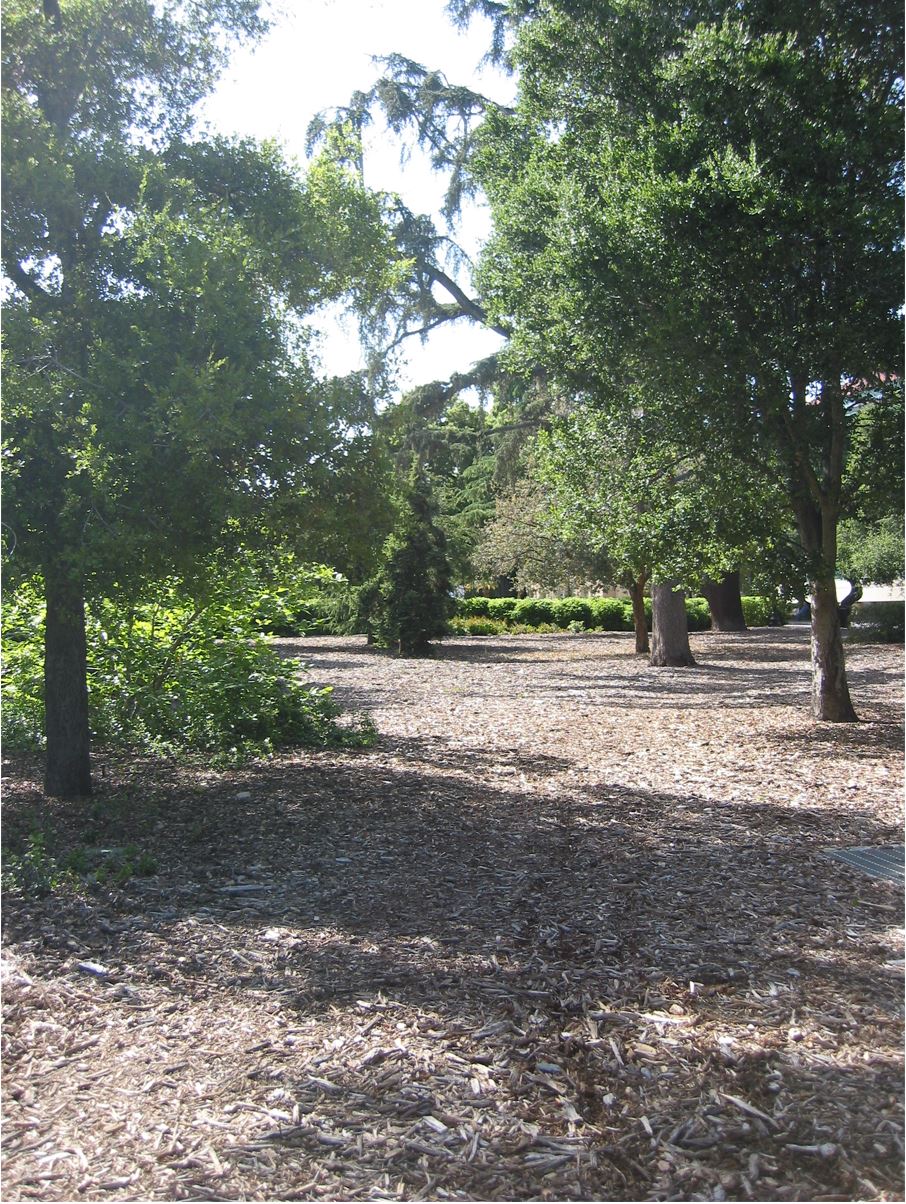}}
 		}\hspace{0.1in} 
		\quad
		\vspace{-0.1in}

		\subfigure{
		    \parbox[c]{2mm}{\rotatebox[origin=c]{90}{\small Ground Truth}}
		}
		\subfigure{
			\parbox[c]{0.1\linewidth}{\includegraphics[width=5em]{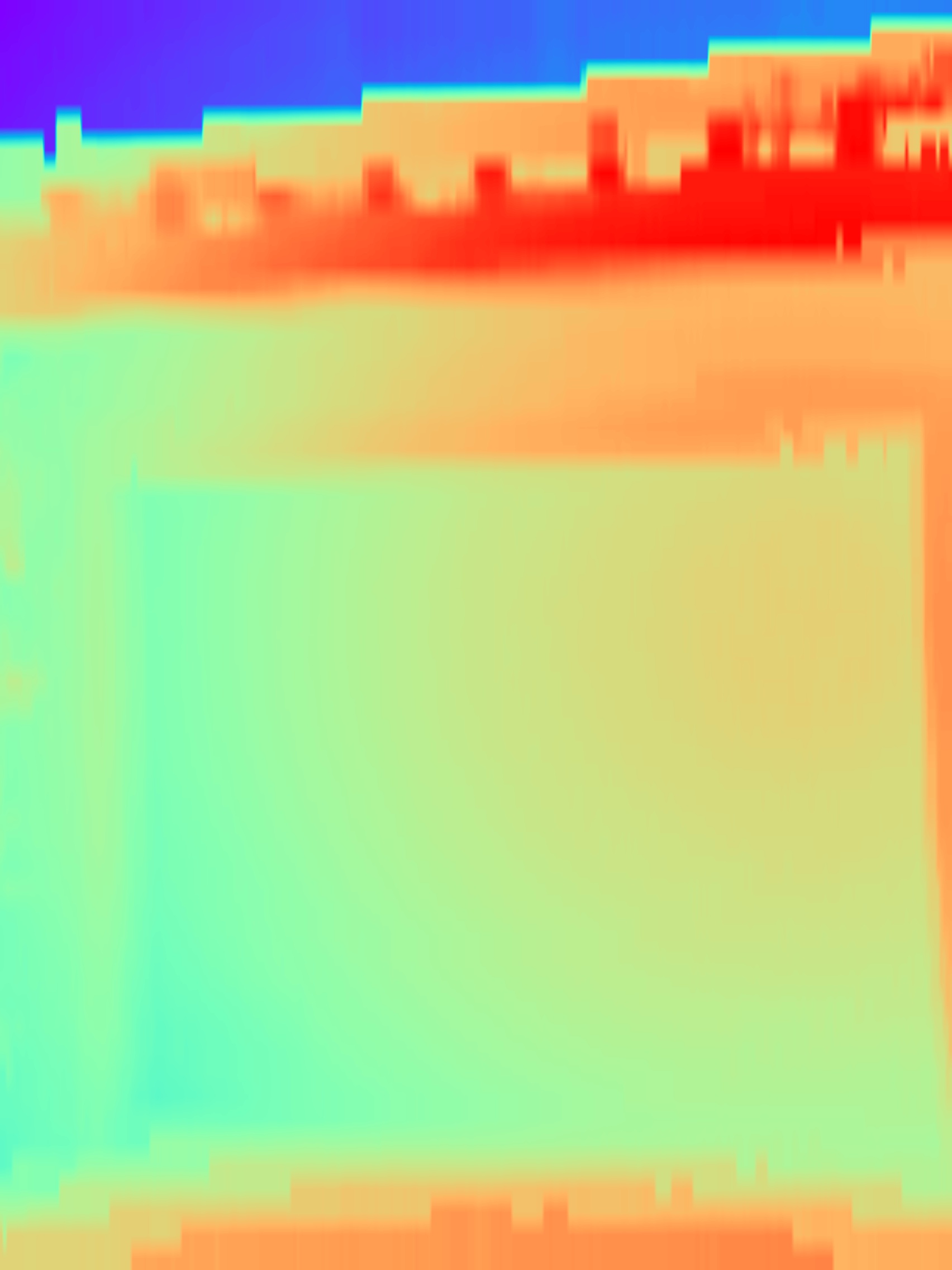}}
		}\hspace{0.1in}
		\quad
 		\subfigure{
 			\parbox[c]{0.1\linewidth}{\includegraphics[width=5em]{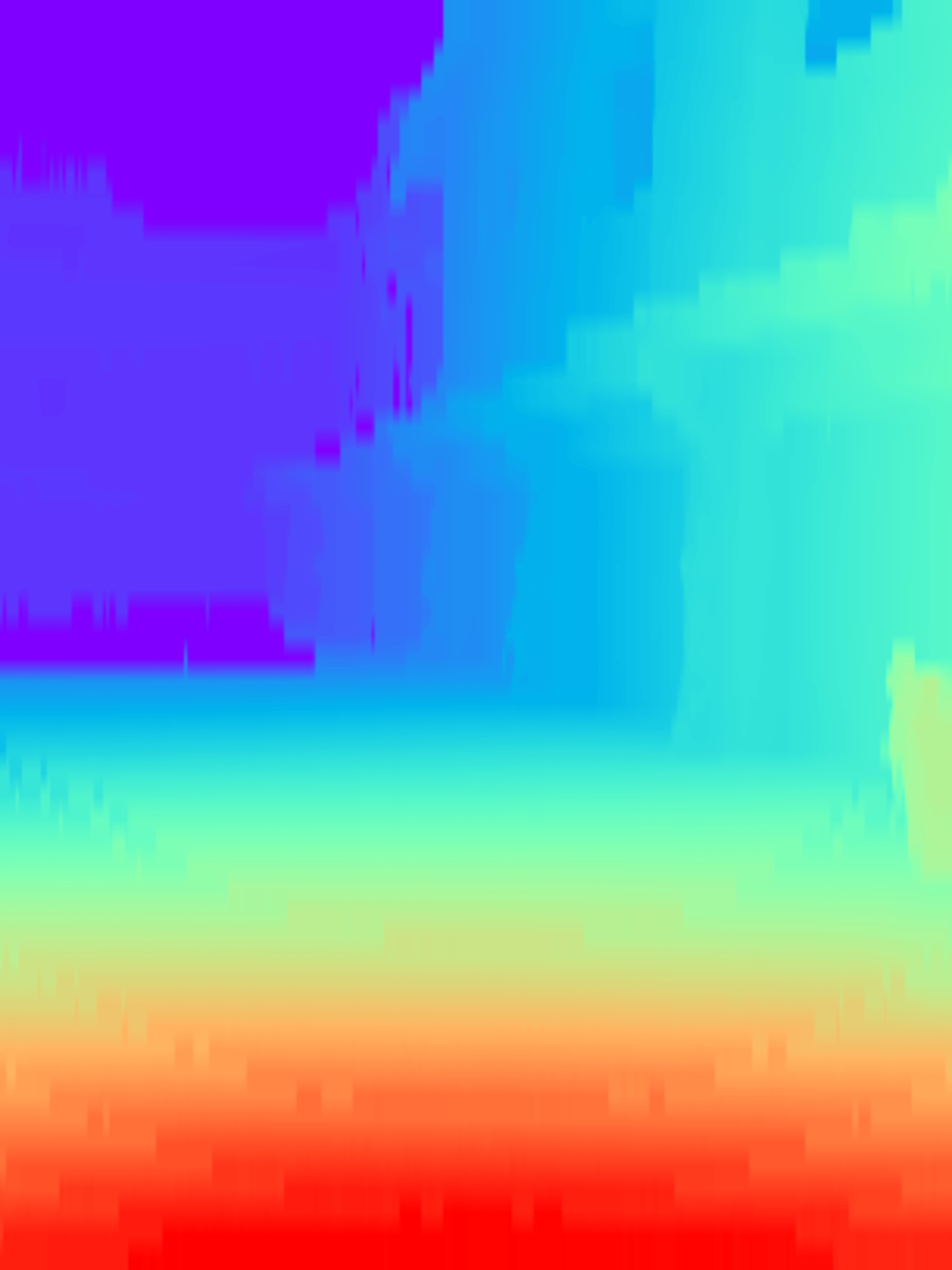}}
 		}\hspace{0.1in} 
 		\quad
 		\subfigure{
 			\parbox[c]{0.1\linewidth}{\includegraphics[width=5em]{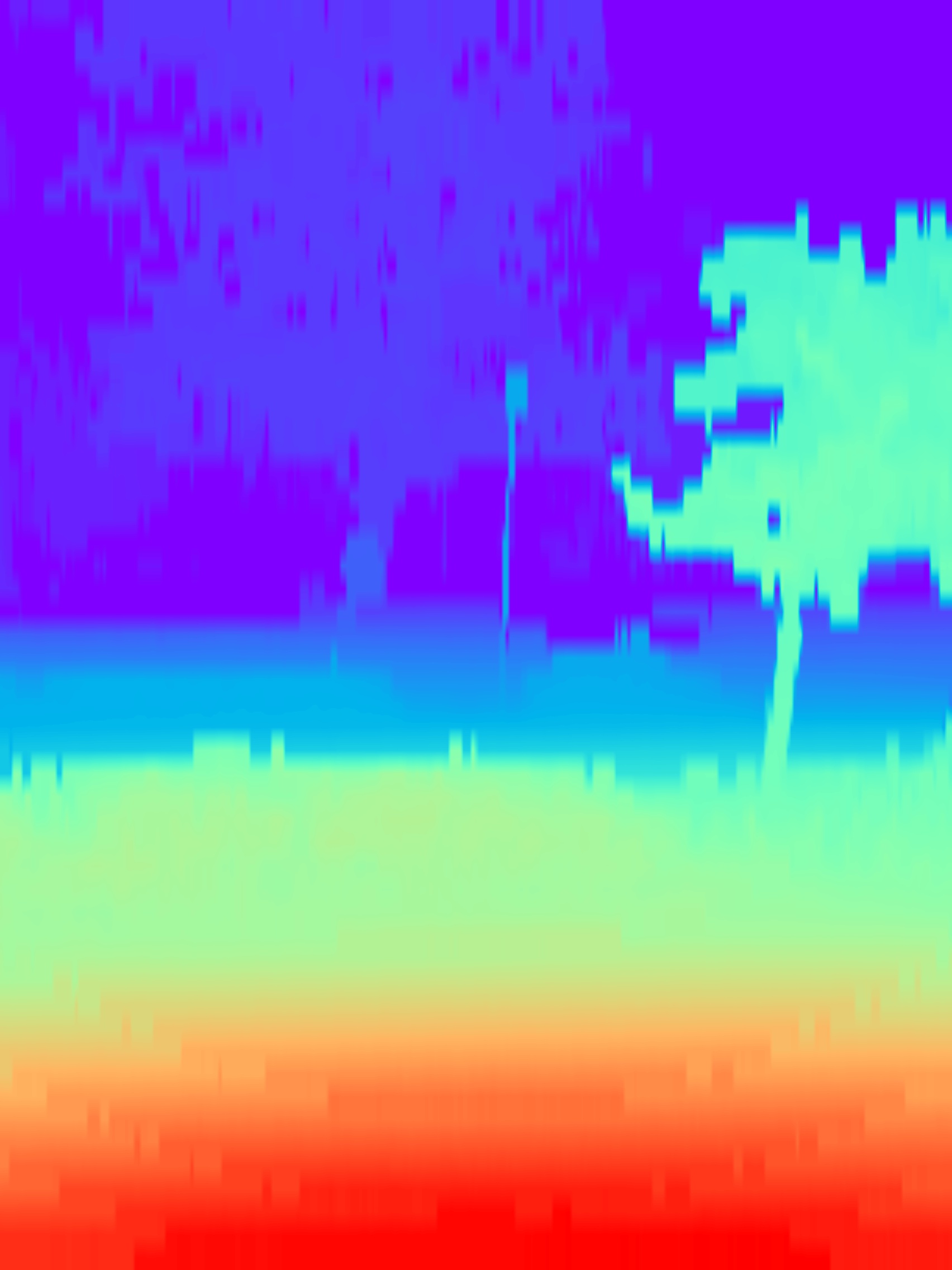}}
 		}\hspace{0.1in} 
		\quad
		\subfigure{
 			\parbox[c]{0.1\linewidth}{\includegraphics[width=5em]{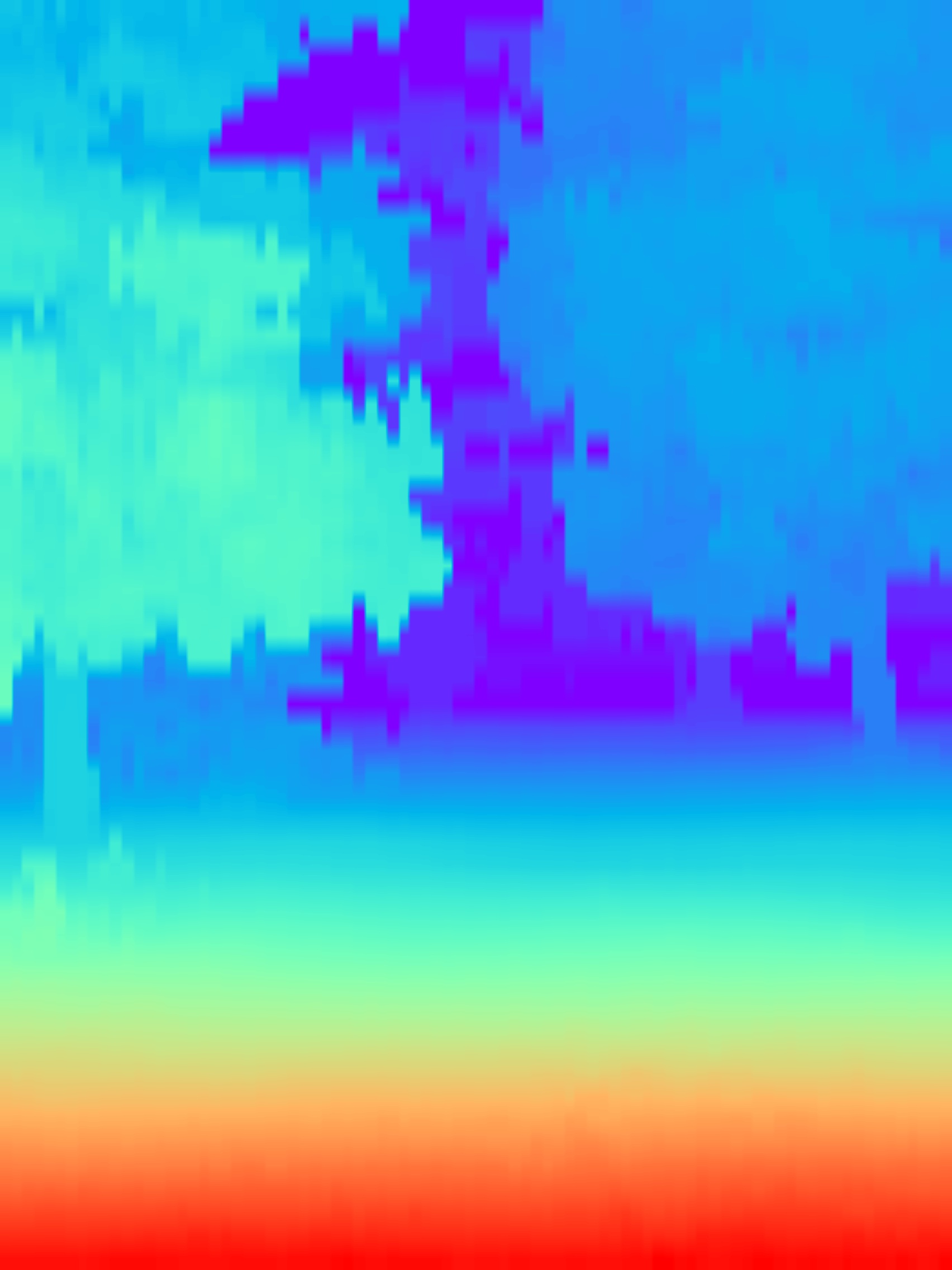}}
 		}\hspace{0.1in} 
		\quad
		\vspace{-0.1in}

		\subfigure{
		    \parbox[c]{2mm}{\rotatebox[origin=c]{90}{\small DDVO \cite{wang2018learning}}}
		}
		\subfigure{
			\parbox[c]{0.1\linewidth}{\includegraphics[width=5em]{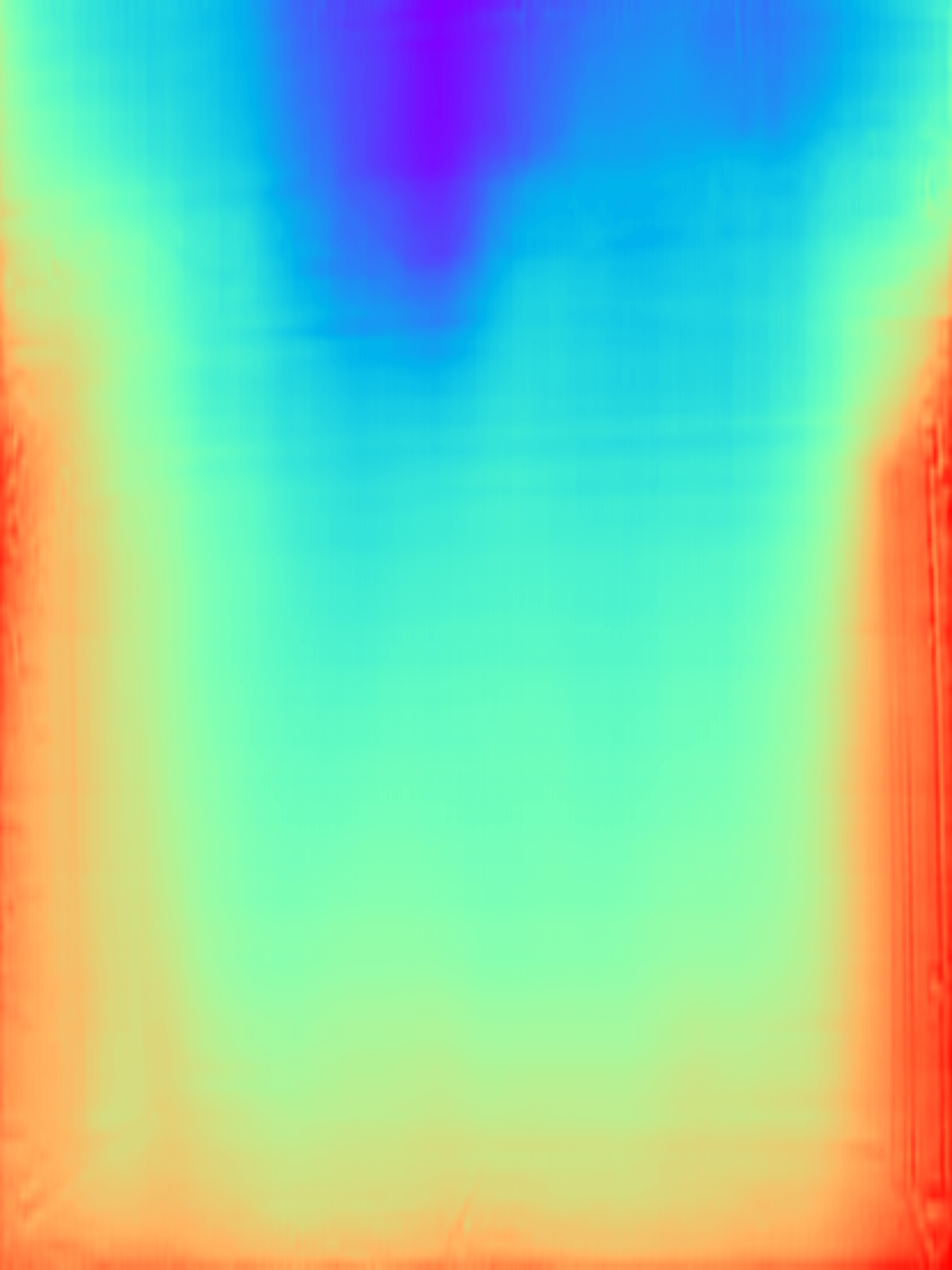}}
		}\hspace{0.1in}
		\quad
 		\subfigure{
 			\parbox[c]{0.1\linewidth}{\includegraphics[width=5em]{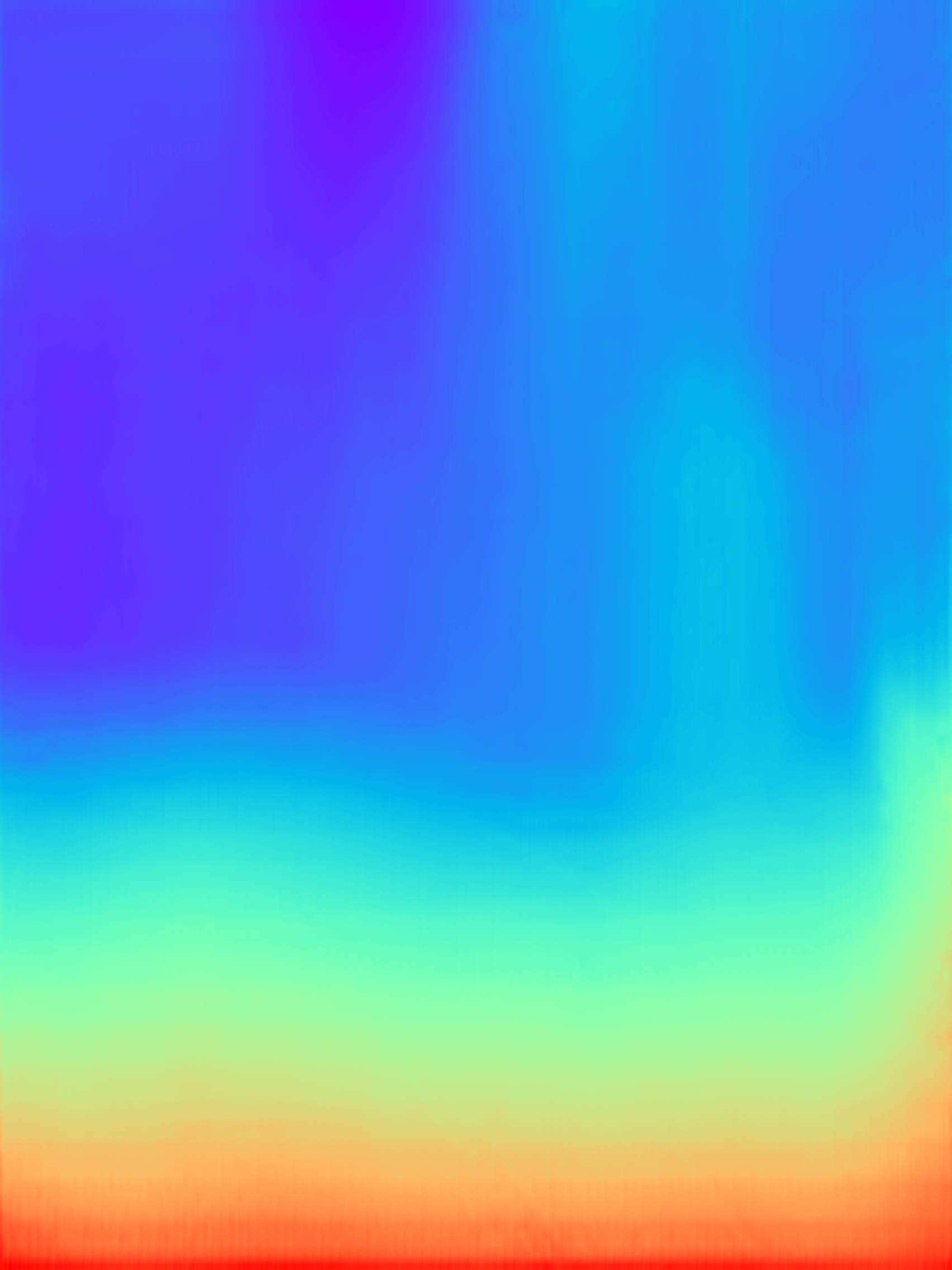}}
 		}\hspace{0.1in} 
 		\quad
 		\subfigure{
 			\parbox[c]{0.1\linewidth}{\includegraphics[width=5em]{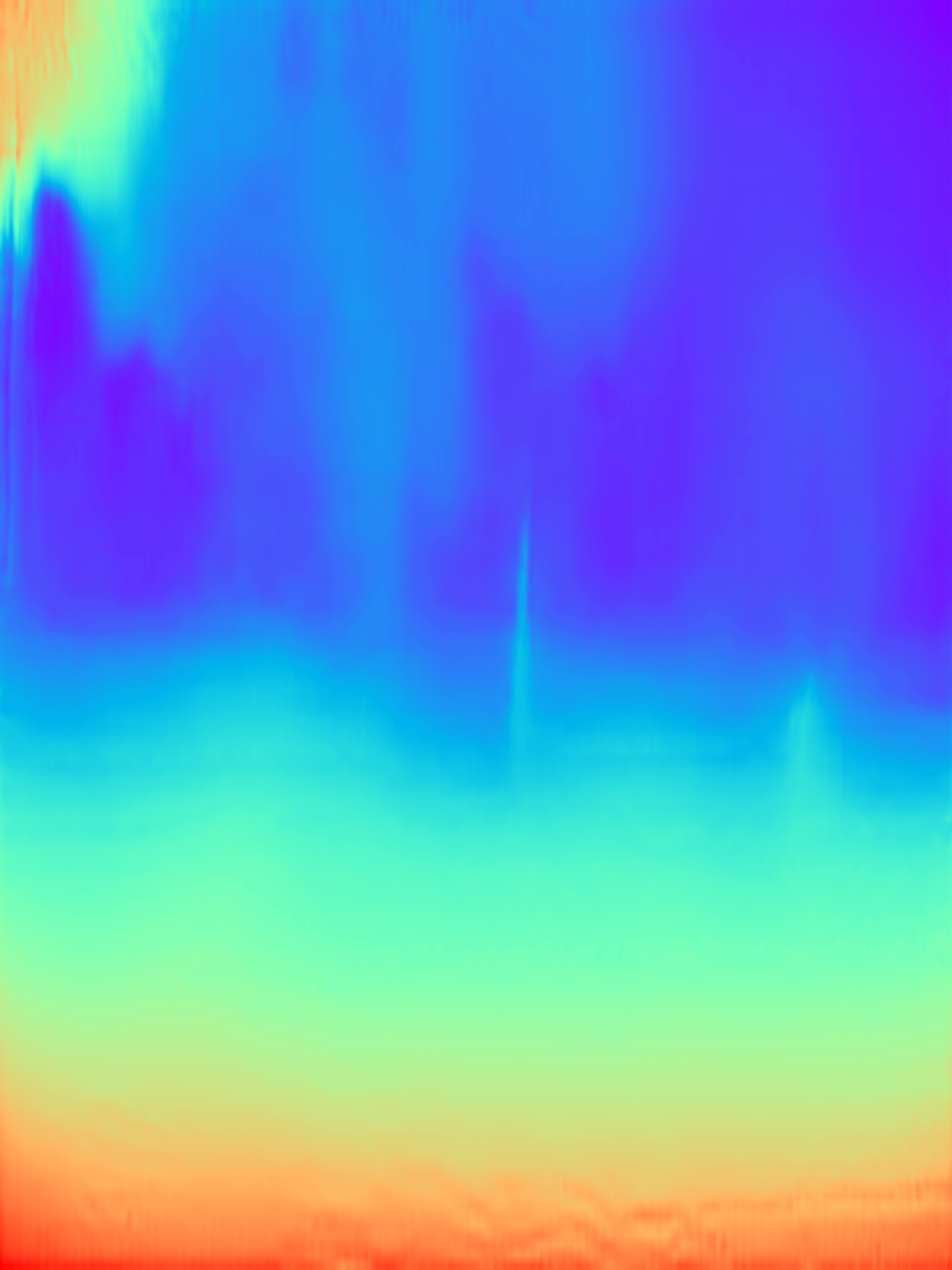}}
 		}\hspace{0.12in} 
		\quad
		\subfigure{
 			\parbox[c]{0.1\linewidth}{\includegraphics[width=5em]{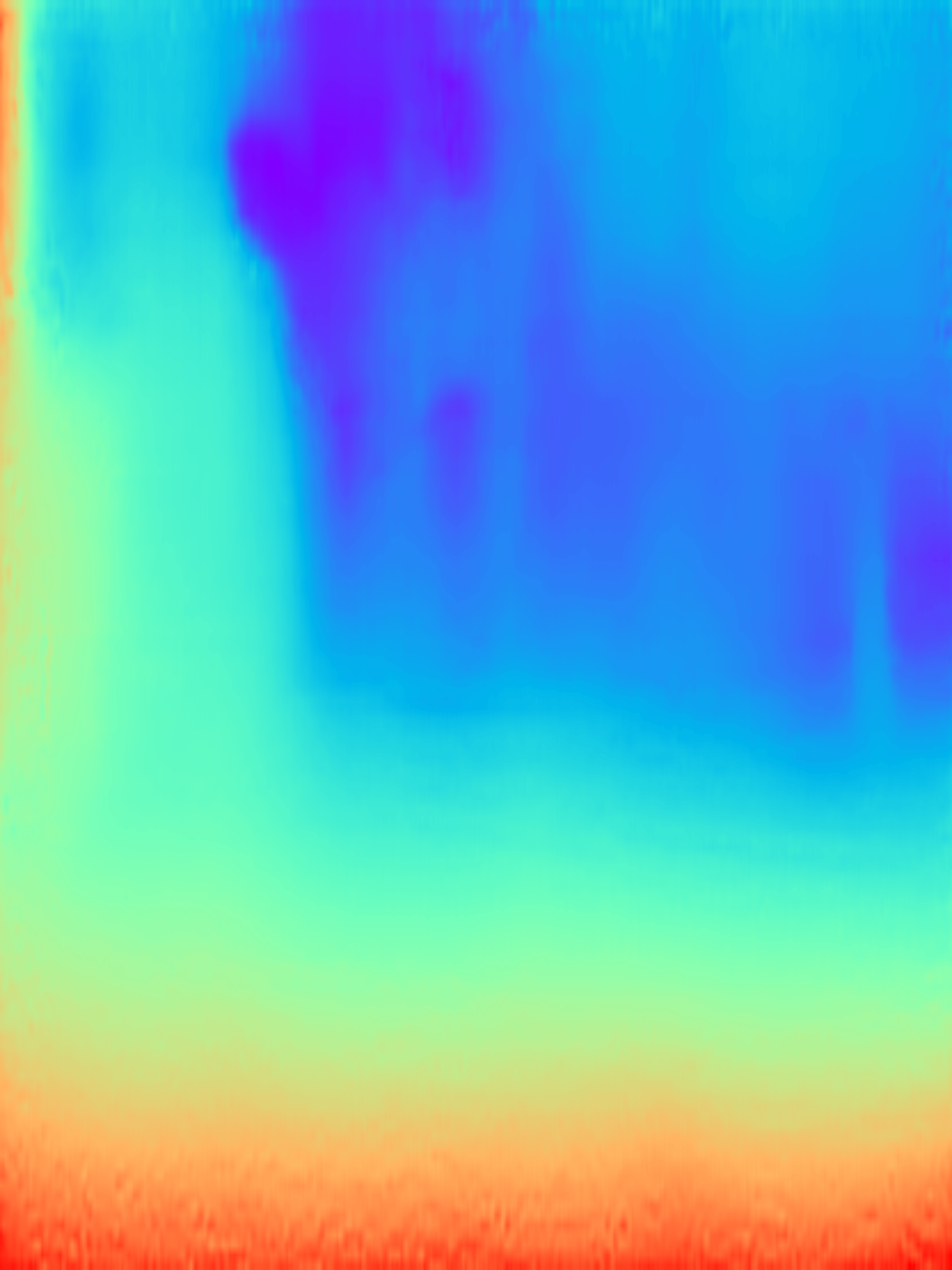}}
 		}\hspace{0.1in} 
		\quad
		\vspace{-0.1in}

		\subfigure{
		    \parbox[c]{2mm}{\rotatebox[origin=c]{90}{\small mono2 \cite{godard2019digging}}}
		}
		\subfigure{
			\parbox[c]{0.1\linewidth}{\includegraphics[width=5em]{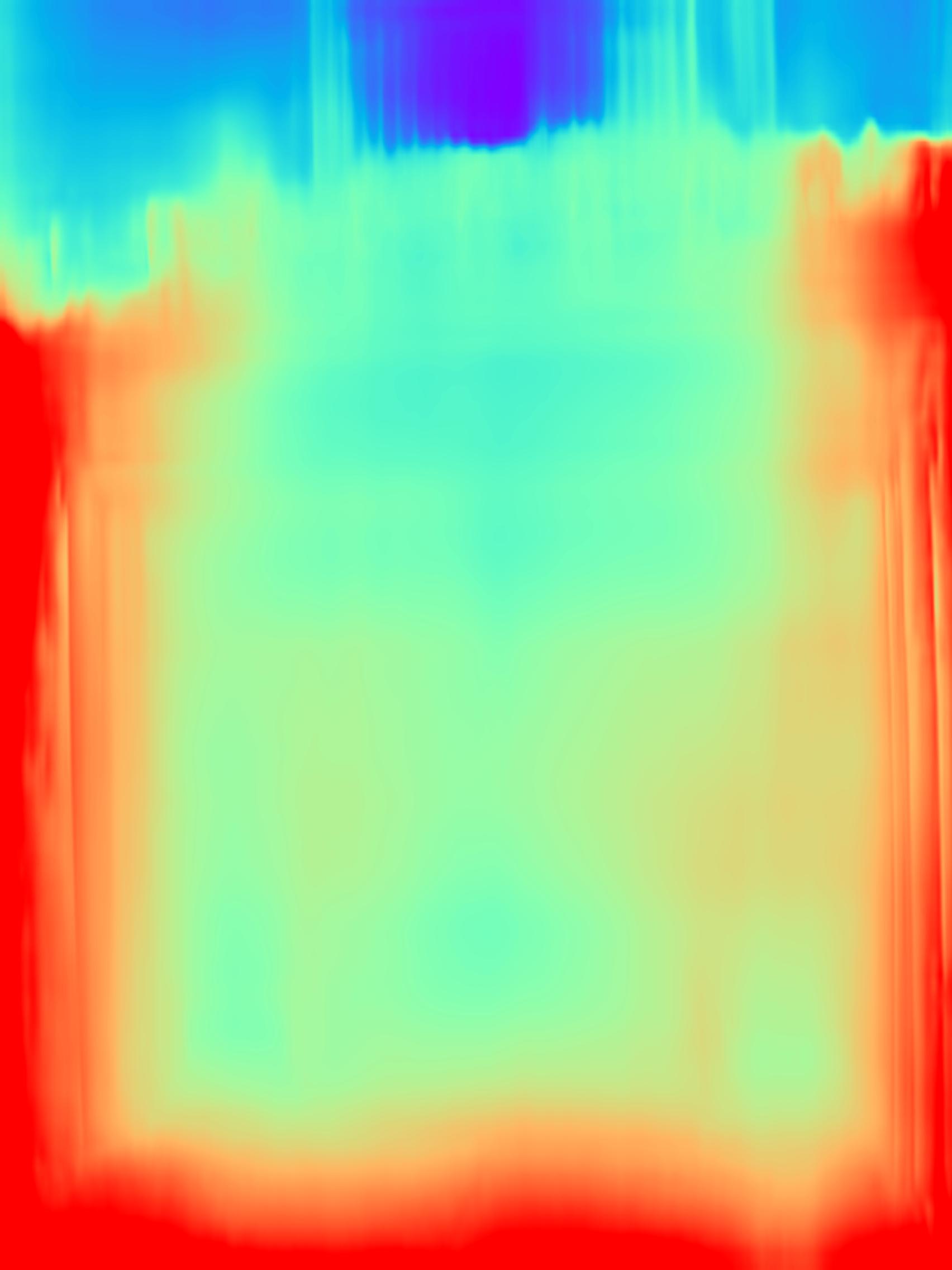}}
		}\hspace{0.1in}
		\quad
 		\subfigure{
 			\parbox[c]{0.1\linewidth}{\includegraphics[width=5em]{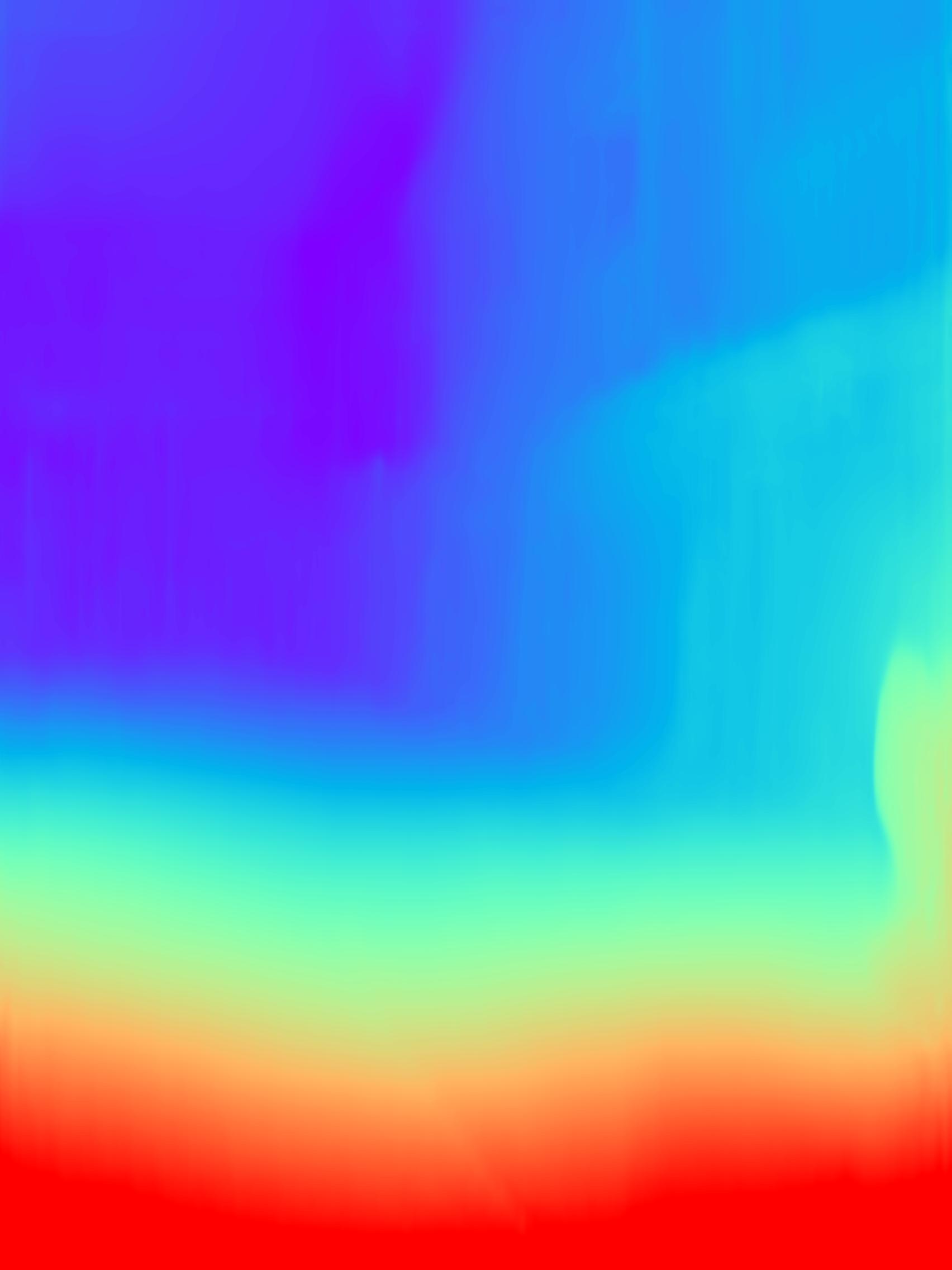}}
 		}\hspace{0.1in} 
 		\quad
 		\subfigure{
 			\parbox[c]{0.1\linewidth}{\includegraphics[width=5em]{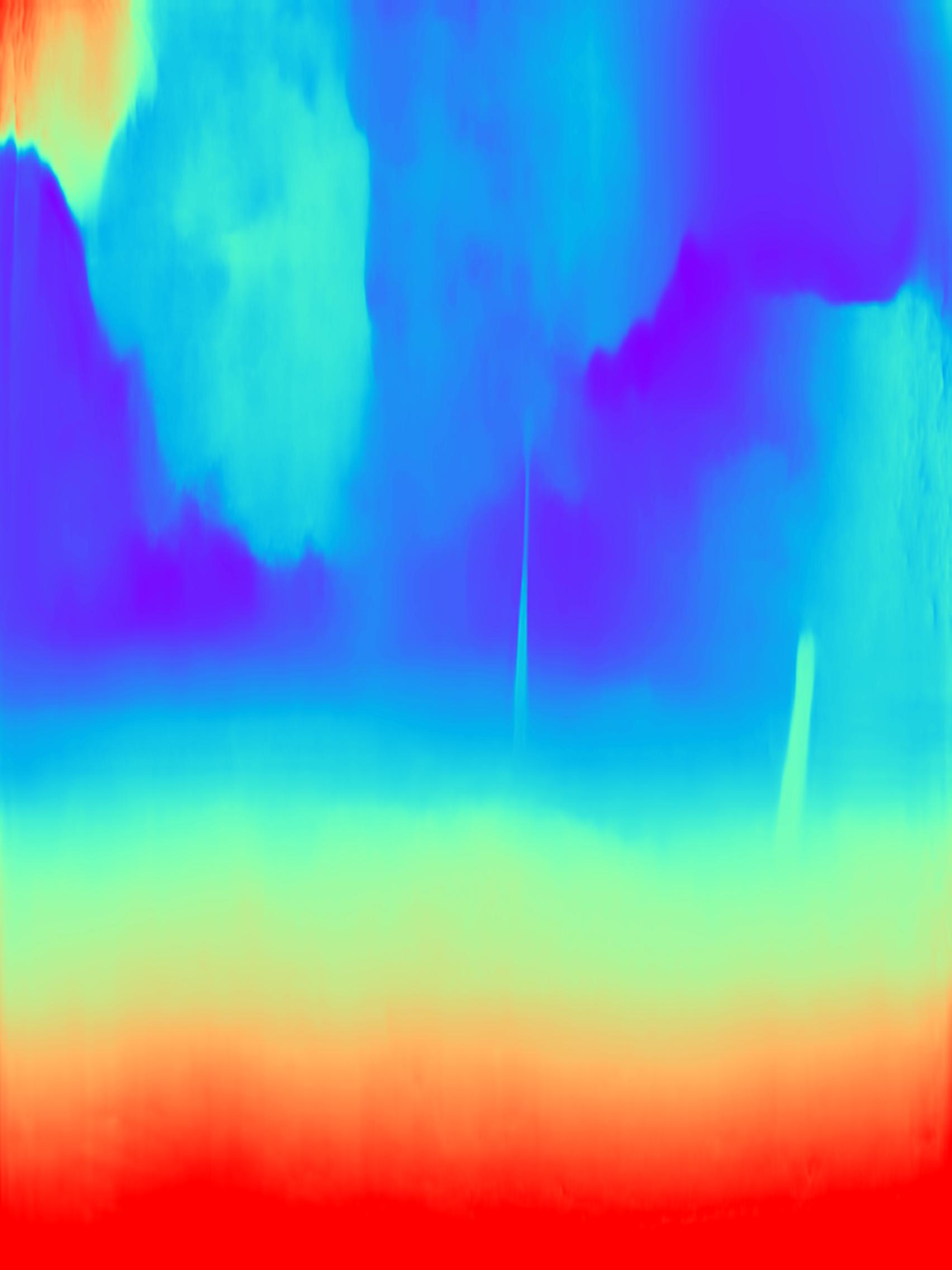}}
 		}\hspace{0.1in} 
		\quad
		\subfigure{
 			\parbox[c]{0.1\linewidth}{\includegraphics[width=5em]{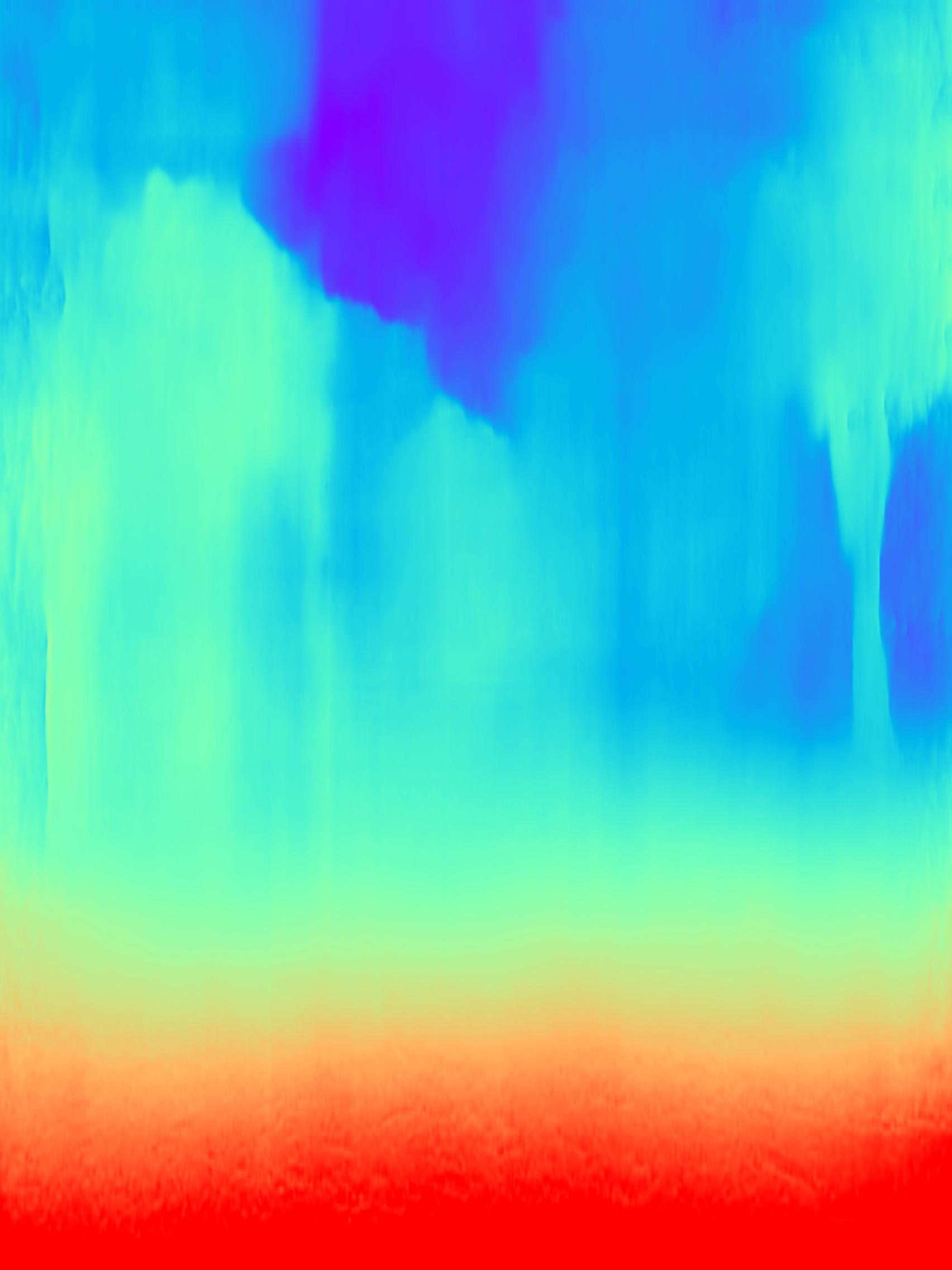}}
 		}\hspace{0.1in} 
		\quad
		\vspace{-0.1in}

		\subfigure{
		    \parbox[c]{2mm}{\rotatebox[origin=c]{90}{Ours}}
		}
		\subfigure{
			\parbox[c]{0.1\linewidth}{\includegraphics[width=5em]{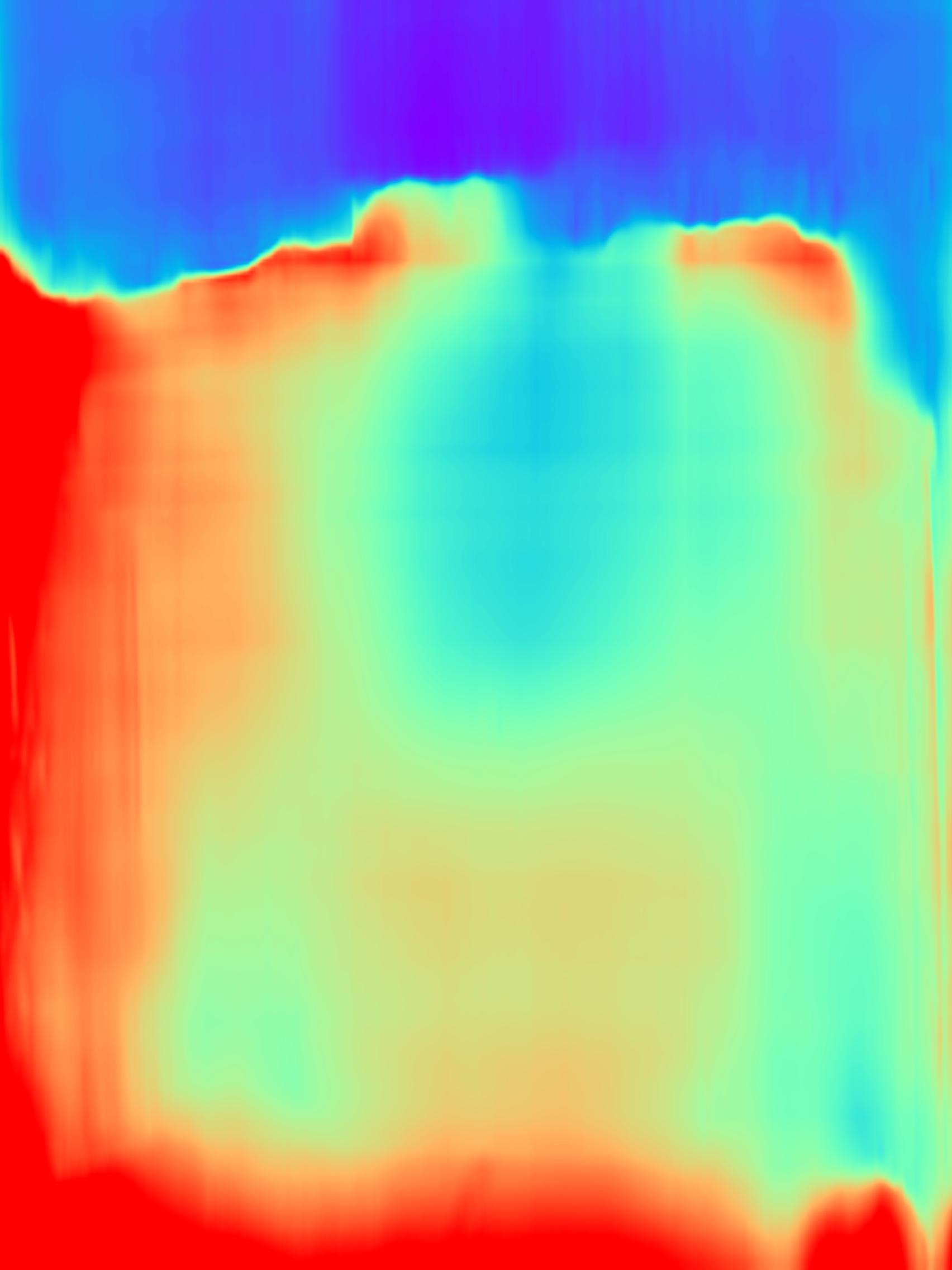}}
		}\hspace{0.1in}
		\quad
 		\subfigure{
 			\parbox[c]{0.1\linewidth}{\includegraphics[width=5em]{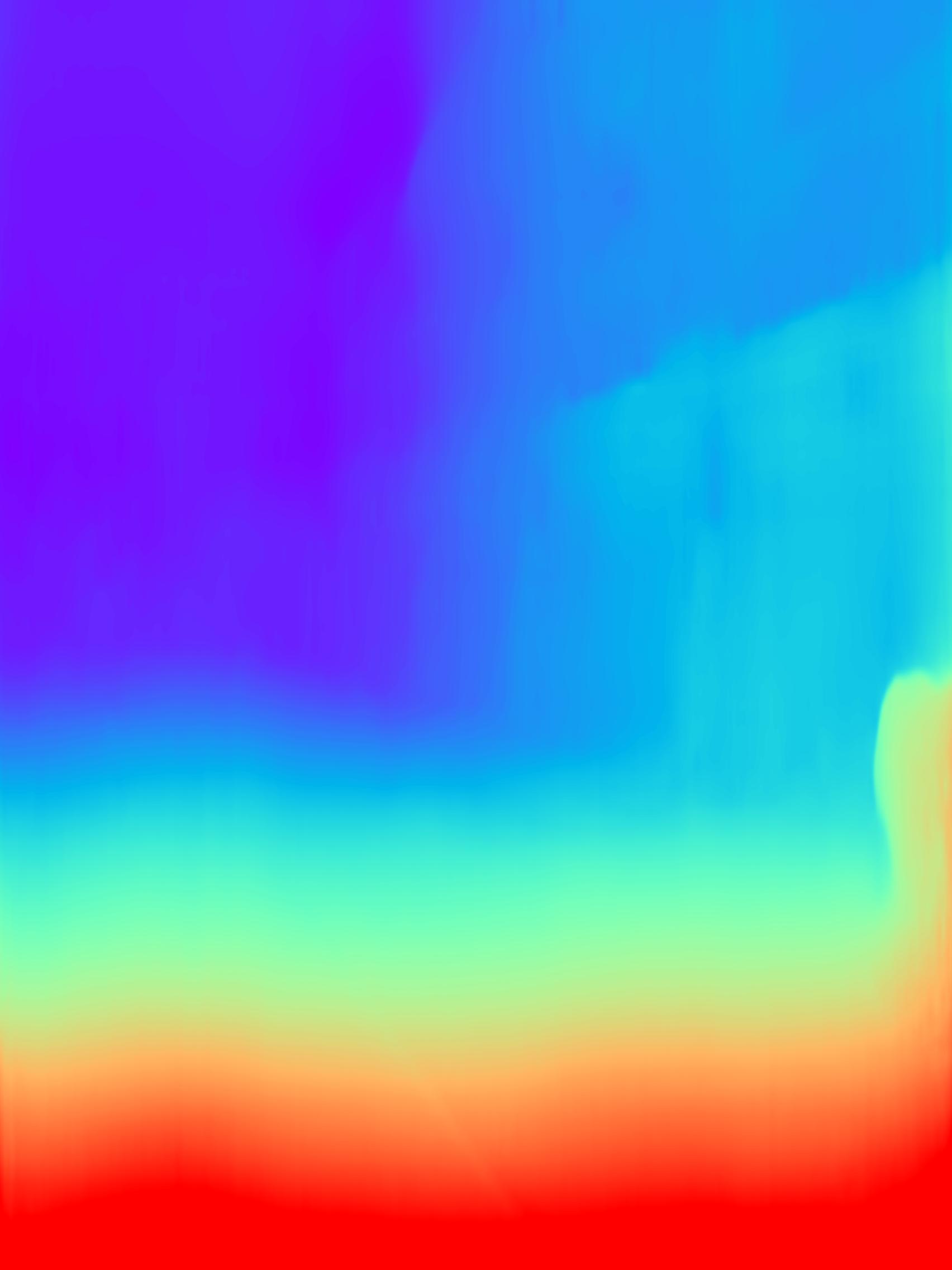}}
 		}\hspace{0.1in} 
 		\quad
 		\subfigure{
 			\parbox[c]{0.1\linewidth}{\includegraphics[width=5em]{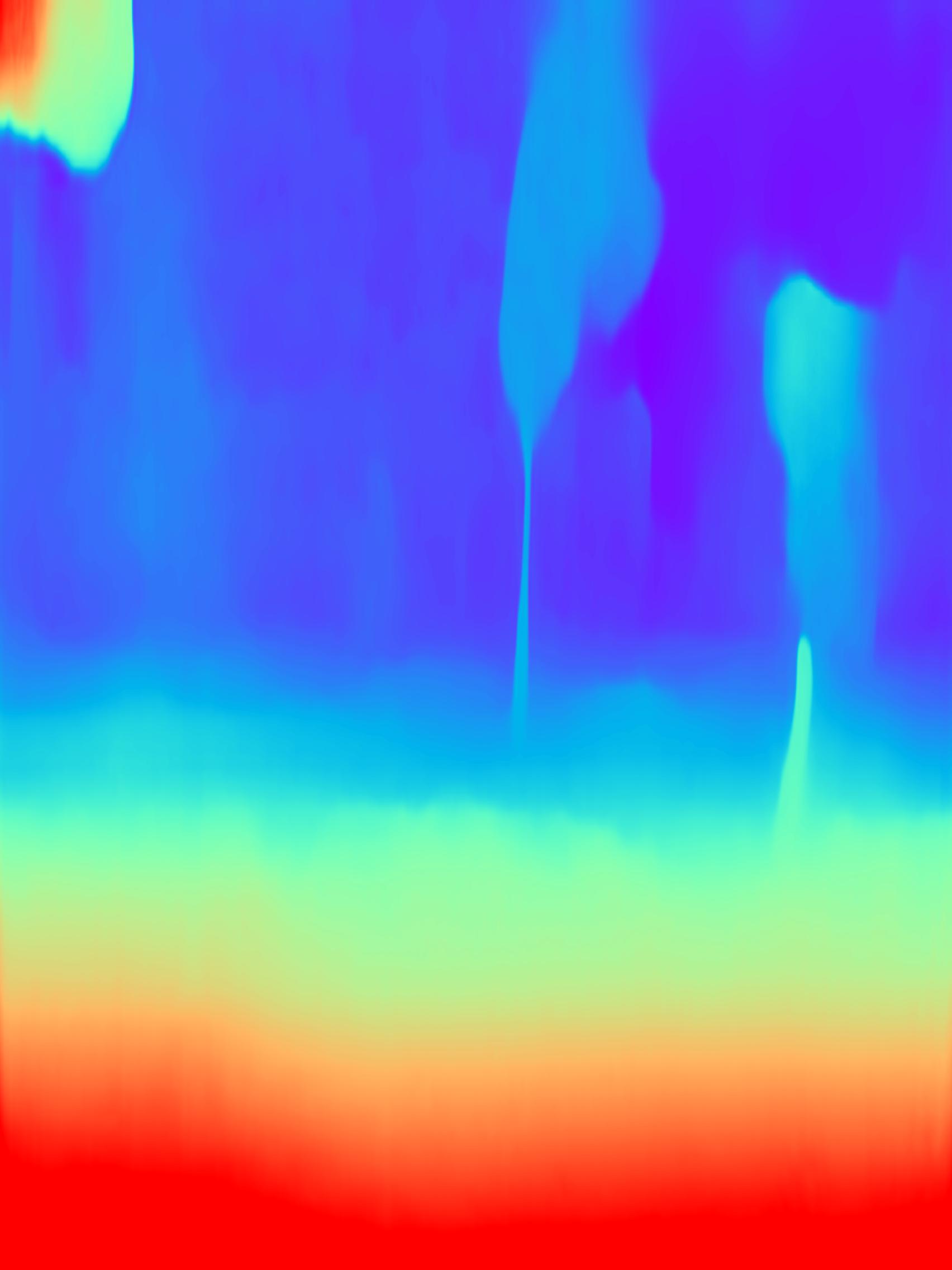}}
 		}\hspace{0.1in} 
		\quad
		\subfigure{
 			\parbox[c]{0.1\linewidth}{\includegraphics[width=5em]{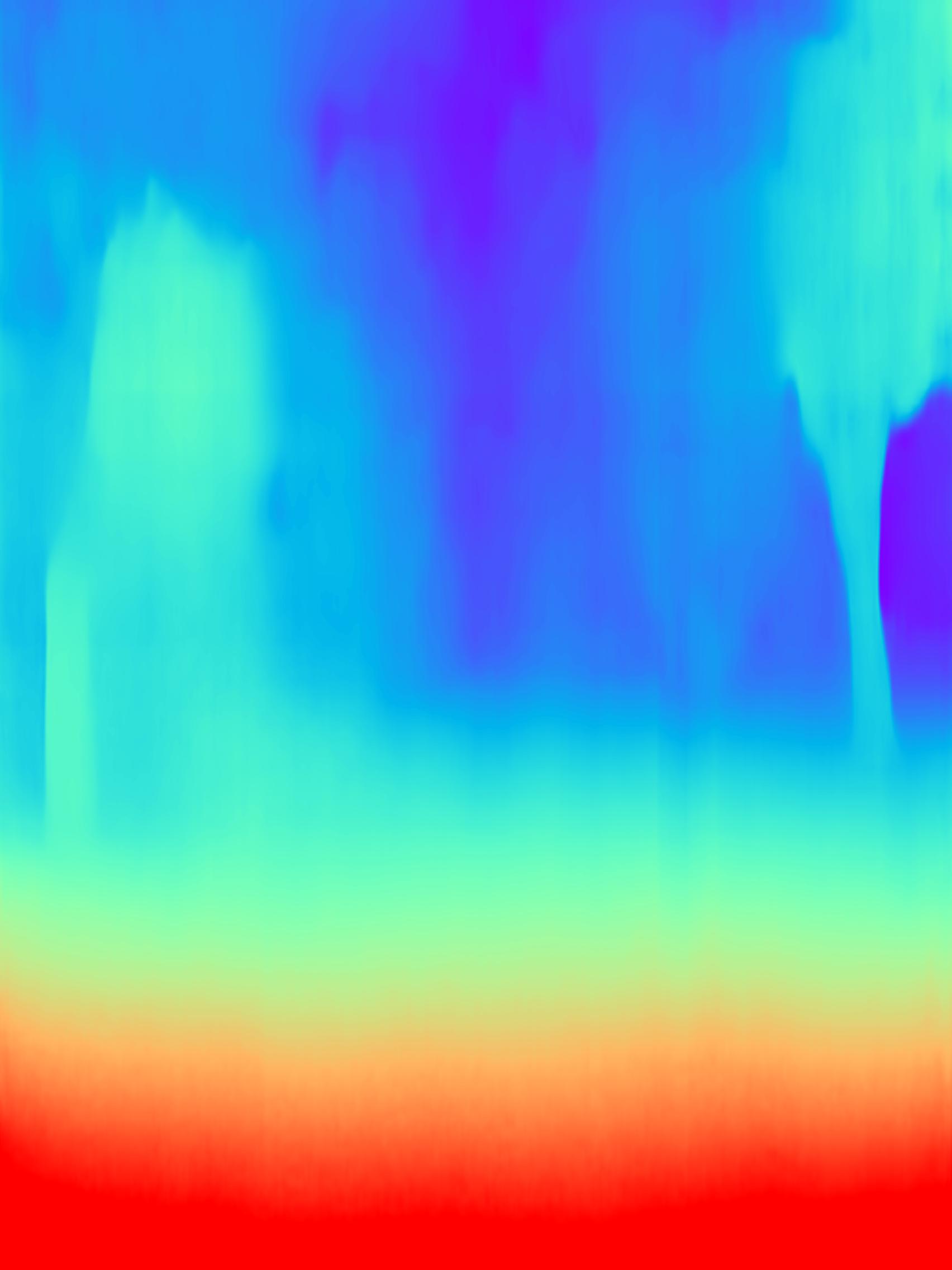}}
 		}\hspace{0.1in} 
		\quad
		\vspace{-0.1in}

		\caption{\textbf{Qualitative Results - Make3D \cite{saxena2008make3d}} }
		\vspace{-8mm}
	\label{make3d visual}
	\end{center}
\end{figure}

\subsection{Results on KITTI dataset \cite{geiger2012we}}
Table ~\ref{KITTI quan} shows the quantitative results for our experimental analysis on the KITTI test split by Eigen \cite{eigen2014depth}.
As evident from the results, our model outperforms other methods in all evaluation metrics.

Figure \ref{KITTI visual} shows the qualitative results for estimating depth on the images from the KITTI test set.
As visible from the depth maps, the results obtained using our approach have depth values which can clearly delineate the objects and do not have edge-blurring effect as in the other approaches.


\begin{table*}[b]
\centering
\setlength{\tabcolsep}{2.5mm}
\renewcommand{\arraystretch}{1.3} 
\begin{tabular}{|c|c||c|c|c|c||c|c|c|}
\hline
\multirow{2}{23mm}{\footnotesize Deformable Support Window } & \multirow{2}{21mm}{\footnotesize Pixel Movement Prediction} & Abs Rel        & Sq Rel         & RMSE           & RMSE $log$    & {\footnotesize $\delta < 1.25$} & {\footnotesize $\delta < 1.25^{2}$} & {\footnotesize $\delta < 1.25^{3}$} \\ \cline{3-9} 
                             &                             & \multicolumn{4}{c||}{Lower is better}                              & \multicolumn{3}{c|}{Higher is better}                       \\ \hline \hline
              &             & 0.179          & 0.254             & 3.578           & 0.131              & 0.527              & 0.801              & 0.753            \\ \hline
\Checkmark    &             & 0.150          & 0.187             & 3.221           & 0.125              & 0.624              & 0.828              & 0.826            \\ \hline
              & \Checkmark  & {\ul 0.064}    & {\ul 0.178}       & {\ul 2.356}     & {\ul 0.104}        & {\ul 0.878}        & {\ul 0.861}        & {\ul 0.865}      \\ \hline
\Checkmark    & \Checkmark  & \textbf{0.049} & \textbf{0.157}    & \textbf{1.831}  & \textbf{0.071}     & \textbf{0.994}     & \textbf{0.996}     & \textbf{0.999}   \\ \hline
\end{tabular}
\vspace{-1mm}
\caption{\textbf{KITTI \cite{geiger2012we} Ablation Study:} Best results for each metric are in \textbf{bold}; second best are \uline{underlined}.}
\vspace{-6mm}
\label{KITTI abl}
\end{table*}


\subsection{Results on Make3D dataset \cite{saxena2008make3d}}
Quantitative results obtained on applying our trained model on the images from the Make3D dataset are reported using Abs Rel, Sq Rel, RMSE, and RMSE $log$ as the metrics.
As shown in Table \ref{make3d qua}, our model performs better than all other approaches.

Figure \ref{make3d visual} shows the qualitative depth estimation results on four images from the Make3D dataset.
Visual comparison demonstrates that our results are closer to the ground truth as compared to those of the other approaches.

\subsection{Results on NYU Depth V2 \cite{silberman2012indoor}} 
As shown in Table \ref{NYU qua}, our model performs better than all other approaches. It gives best values for the Abs Rel metric, Sq Rel metric, and second to the best values for the remaining two metrics RMSE, and RMSE $log$. The results on NYU Depth V2 shows our model not only can work in an outdoor environment but also in an indoor scenario.

\subsection{Ablation Study on KITTI dataset}
We conduct an ablation study on the KITTI dataset in order to better understand the importance of the two proposed modules - deformable support window and pixel movement prediction.
Table \ref{KITTI abl} shows the results of the different variants of our model on the KITTI dataset using the Eigen split.
 As seen in the results, the baseline model with just our neural network architecture but none of the two proposed modules, performs the worst.
Whereas the model with all our proposed contributions performs the best with significant improvements.
The variant with the pixel movement prediction module works the second best.
This proposed 3-part pixel movement prediction along with the pixel movement triangle constraint loss is the primary contributor towards successful monocular depth estimation.
The deformable support window module also contributes to improving the quantitative results.
However, this module's importance is realized more from a qualitative standpoint of obtained non-blurred edges.

\section{Conclusion}\label{con}
In this paper, we propose a novel deep neural network called PMPNet for monocular depth estimation in dynamic scenes.
The model performs individual pixel movement prediction so as to lift the static scene assumption while simultaneously ensuring geometric consistency during depth estimation.
We assume that objects move along a straight line and use this to predict two possible movements for each pixel and one straight line to constrain them.
A novel pixel movement triangle constraint loss is proposed for restricting the relationship between the pixel movement predictions and the straight line, thereby overcoming the inconsistency between the pixel movement and the depth map.
The proposed model also consists of a support window module consisting of deformable convolutions which overcomes the incorrect depth estimation problem around object edges.
Experiments were conducted on two outdoor datasets - KITTI and Make3D, and one indoor dataset - NYU Depth V2.
Results of these, alongside the ablation study results on the KITTI dataset, demonstrate the effectiveness of our proposed PMPNet for monocular depth estimation in dynamic scenes.

\bibliographystyle{IEEEtran}
\bibliography{egbib}
%



\end{document}